\documentclass[preprint,12pt]{elsarticle}

\usepackage{amssymb}
\usepackage{amsmath}
\usepackage{graphicx}
\usepackage{subcaption}
\usepackage{hyperref}
\usepackage{soul}
\usepackage[table]{xcolor}
\usepackage{algorithm}
\usepackage{caption}
\usepackage{float}

\usepackage{algpseudocode}

\DeclareMathOperator*{\argmax}{arg\,max}

\newcommand{\ex}{\mathbb{E}}

\journal{Engineering Application of Artificial Intelligence}

\begin{document}

\begin{frontmatter}



\title{A Study of the Efficacy of Generative Flow Networks for Robotics and Machine Fault-Adaptation}


\author[inst1]{Zahin Sufiyan}
\author[inst1,inst4]{Shadan Golestan}
\affiliation[inst1]{organization={Department of Computing Science, University of Alberta},
            addressline={8900 114 St NW}, 
            city={Edmonton},
            postcode={T6G 2S4}, 
            state={Alberta},
            country={Canada}}

\author[inst2]{Shotaro Miwa}
\affiliation[inst2]{organization={Advanced Technology R\&D Center, Mitsubishi Electric Corporation},
            addressline={8-1-1 Tsukaguchi-honmachi, Amagasaki-shi}, 
            city={Hyogo},
            postcode={661-8661}, 
            country={Japan}}

\author[inst3]{Yoshihiro Mitsuka}
\affiliation[inst3]{organization={Information Technology R\&D Center, Mitsubishi Electric Corporation},
            addressline={5-1-1 Ofuna}, 
            city={Kamakura-shi},
            state={Kanagawa},
            postcode={247-8501}, 
            country={Japan}}

\author[inst1,inst4]{Osmar Zaiane}
\affiliation[inst4]{organization={Alberta Machine Intelligence Institute},
            addressline={10065 Jasper Ave \#1101}, 
            city={Edmonton},
            postcode={T5J 3B1}, 
            state={Alberta},
            country={Canada}}

\begin{abstract}

Advancements in robotics have opened possibilities to automate tasks in various fields such as manufacturing, emergency response and healthcare. 
However, a significant challenge that prevents robots from operating in real-world environments effectively is out-of-distribution (OOD) situations, wherein robots encounter unforseen situations. One major OOD situations is when robots encounter faults,
making fault adaptation essential for real-world operation for robots.
Current state-of-the-art reinforcement learning algorithms show promising results but suffer from sample inefficiency, leading to low adaptation speed due to their limited ability to generalize to OOD situations.
Our research is a step towards adding hardware fault tolerance and fast fault adaptability to
machines. In this research, our primary focus is to investigate the efficacy of
generative flow networks in robotic environments,
particularly in the domain of machine fault adaptation. 
We simulated a robotic environment called Reacher in our experiments. 
We modify this environment to introduce four distinct fault environments
that replicate real-world machines/robot malfunctions.
The empirical evaluation of this research indicates that continuous generative flow networks (CFlowNets) indeed have the capability to add adaptive behaviors in machines under adversarial conditions. Furthermore, the comparative analysis of CFlowNets with reinforcement learning algorithms also provides some key insights into the performance in terms of adaptation speed and sample efficiency. Additionally, a separate study investigates the implications of transferring knowledge from pre-fault task to post-fault environments. 
Our experiments confirm that CFlowNets has the potential to be deployed in a real-world machine and it can demonstrate adaptability in case of malfunctions to maintain functionality. 
%

\end{abstract}




\end{frontmatter}



\section{Introduction}
\label{sec:intro}
Advancements in robotics have revolutionized industries,
and has numerous benefits to society,
with various applications ranging from 
manufacturing~\cite{9911168}, search and rescue operations~\cite{9629248}, disaster response~\cite{7992658}, to healthcare~\cite{7314984}. 
Remarkably, they are indispensable tools for scientific research, 
enabling the exploration of remote areas such as other planets~\cite{papadakis2013terrain}, and underwater areas~\cite{liu2023plant}
where sending humans might be dangerous, extremely costly, or infeasible.
These robotic systems rely on a variety of technologies, 
particularly artificial intelligence,
to operate efficiently and safely in those environments.
Nevertheless, a significant hurdle to their widespread implementation in unknown environments 
is their vulnerability to out-of-distribution (OOD) situations during deployment~\cite{chen2023adapt}, 
such as unforeseen environmental changes and events such as faults. 
Even in controlled environments such as manufacturing industries,
79.6\% of machine tool downtime is due to factory machines/robots facing OOD situations (especially faults) 
that hinder their operations~\cite{fernandes2022machine,teti2010advanced}. 


To minimize the risk of damage, downtime and ensure the longevity of robots, 
it is essential to develop strategies for fault diagnosis, fault detection, identification, and adaptation.
Several studies have explored various methodologies for fault detection and fault diagnosis \cite{riazi2019detecting,castellano2023multidimensional,chatti2016model}.
These works propose utilizing sensors and monitoring systems to collect data on the performance of the machine, 
which in turn helps with the identification of deviation from normal operating conditions or 
the detection of anomalies such as hardware failure.
Detection and diagnosis mechanisms of this type are only concerned with detecting faults as soon as they arise
but do not address the issue of post-fault adaptation.


To succeed in real-world deployments, especially in remote areas,
robots should be able to adapt their behavior in response to faults.
A traditional approach for adding fault tolerance in machines is through hardware redundancy~\cite{guiochet2017safety,9289420}. 
Hardware redundancy refers to the duplication of essential machine/robot components 
in order to replace similar broken components whenever a fault occurs. 
Although effective, the duplication of components increases the size, weight, power consumption, and financial cost of the machine. 
Additionally, retrofitting redundant components into existing machines can also be challenging 
since the original blueprint of the machine may not allow such modifications. Using more materials to build machines with redundant components also has a negative impact on the environment. 
As a result, the problem of machine adaptation has gradually shifted from hardware modifications, 
to software adaptation, such as learning methods~\cite{9655714,10305254}.
An approach to making machines fault-tolerant that has received considerable attention in recent times is to add the capability of adaptability \cite{9655714}, \cite{10305254}. 
Like living organisms, machines require the ability to adapt to unexpected OOD situations in their environment, 
whether that means adjusting to damage or finding a new way to complete a task. 
For instance, birds adapts to wing injury by changing the way it flaps its wings. Machines can also take inspiration from nature by adopting a fault-tolerant approach, similar to how some animals and plants are able to survive and thrive in hostile environments. 
In fact, many researchers and engineers are now turning to biomimicry, the process of imitating nature’s designs and processes, to create machines that are better able to handle unexpected challenges \cite{theengineer_forces_of_nature}. 
The approach entails building machines that are capable of recognizing and adjusting to variations in the environment, including the presence of faults, to ensure continued operation.

There is a rich and growing body of literature that frame the problem of machine fault adaptation
as a Reinforcement Learning (RL) problem~\cite{chen2023adapt,cully2015robots,song2020rapidly},
wherein an RL agent attempts to learn specific behaviours via interaction with an environment,
and generalize the behaviours to unseen variations of the task~\cite{sutton2018reinforcement}.
Due to the high-dimensional nature of the robotic tasks~\cite{arulkumaran2017brief,gu2016deep},
these studies primarily focused on deep RL algorithms
for their ability to handle such complexities. 
However, a significant challenge associated with deep RL algorithms 
is their sample inefficiency~\cite{raziei2021adaptable},
as these algorithms require large training datasets, 
which can be difficult or expensive to collect for complex and specific robotic tasks, 
especially for fault adaptation.
Another challenge of RL algorithms lies in the exploration and exploitation trade-off for large-scale problems.
Low exploration can lead to finding sub-optimal solutions,
while promoting high exploration rates at the initial stages of training 
can lead to slower convergence on the optimal policy and 
introduce significant performance variability.

In contrast, Generative Flow Networks (GFlowNets)~\cite{bengio2021gflownet}, 
or its variation for Continuous problems, i.e., Continuous Flow Networks (CFlowNets)~\cite{li2023cflownets},
enhance the exploration capabilities by generating a distribution proportional to rewards over terminating states \cite{bengio2021gflownet}.
CFlowNets have been shown to effectively balance exploration and exploitation,
allowing them to find optimal solutions while remaining generalized to various unseen situations~\cite{zhang2023robust,shen2023towards}. One of the key differences between CFlowNets and reinforcement learning is that CFlowNets generate a distribution over all possible paths and sample from the most rewarding paths with a higher probability which makes them theoretically more sample-efficient than standard reinforcement learning algorithms because they tend to stick to the most rewarding path which can be a local optimum. Even if a high exploration rate is promoted at the initial stages of training, it causes delayed learning of optimal policy and also introduces significant variability in performance. 
Therefore, we hypothesize that this sampling strategy employed by CFlowNets, 
makes them well-suited for robotic applications, as CFlowNets can potentially learn high-dimensional tasks quickly,
and adapt more effectively to OOD situations.
The application of CFlowNets in the context of robotic tasks, 
especially for fault adaptation, 
has not been explored.

In this study, we investigate the application of CFlowNets in the context of continuous exploratory robotic tasks, 
with a particular focus on  fault adaptation as an example of encountering OOD situations. 
To assess the performance of flow networks, 
we compare CFlowNets with state-of-the-art RL algorithms which are DDPG (Deep Deterministic Policy Gradient) \cite{lillicrap2015continuous}, TD3 (Twin-delayed DDPG) \cite{fujimoto2018addressing}, Proximal Policy Optimization (PPO) \cite{schulman2017proximal} and Soft-Actor-Critic (SAC) \cite{haarnoja2018soft} and evaluate their ability to add fault tolerance compared to the flow networks.
Our contributions are:
1) We cast the problem of learning robotic tasks and machine fault adaptation to CFlowNet, 
2) We compare the performance, adaptation speed, and efficacy of CFlowNet with state-of-the-art RL algorithms in machine fault adaptation,
3) We empirically show that CFlowNet achieves significantly faster adaptation with comparable asymptotic performance,
4) We explore different transfer of task knowledge options to determine the best option that enhances CFlowNet's ability to adapt to faults.

To the best of our knowledge, this is the first study that has implemented and evaluated the performance, adaptation speed, and efficacy of generative flow networks in machine fault adaptation. 
As part of the ongoing research on developing more efficient and reliable industrial systems, 
this study provides insights into the potential of the generative flow networks 
as novel frameworks for machine fault adaptation.




\section{Background}
\label{background}
In this section, we provide background information on the contributions of this research. 
Since this study works as a comparative analysis of RL and GFlowNets/CFlowNets, 
we begin by discussing the fundamental concepts of RL in Section~\ref{RL}. 
Afterward, in Section~\ref{gflownets_background}, we delve into the detailed background analysis of GFlowNets/CFlowNets.

\subsection{Reinforcement Learning}
\label{RL}
Reinforcement Learning (RL)~\cite{sutton2018reinforcement} is a type of machine-learning algorithm
where an agent learns to make decisions by taking actions in an environment to achieve a goal. 
In RL, agents are trained to make sequential decisions 
in a particular environment modelled as a Markov Decision Process (MDP), denoted as $
\mathcal{M}{=}(\mathcal{S}, \mathcal{A}, \mathcal{P}, \mathcal{R}, \gamma)$,
where $\mathcal{S}$ is the state space;
$\mathcal{A}$ is the agent's action space;
$\mathcal{P}$ is the environment's transition dynamics;
$\mathcal{R}$ is the reward function; and
$\gamma{\in}[0,1]$ is the discount factor.
At each time $t{=}\{0,1,2,...,h\}$, $h$ being the horizon, 
an agent interacts with an MDP via allowed actions $\mathcal{A}$, 
and gets feedback via the reward function $\mathcal{R}$. 
The state of the environment changes from $s_t{\in}\mathcal{S}$ to $s_{t+1}{\in}\mathcal{S}$ 
based on $\mathcal{P}(s_{t+1}|s_t, a_t)$, 
with $a_t{\in}\mathcal{A}$ being the action taken by the agent at time $t$.
The objective of the agent is to find a policy $\pi$ to maximize an expected discounted sum of rewards 
$J(\pi){=}\ex_{\tau\sim\pi}[\sum_{t=0}^h {\gamma^t \mathcal{R}(s_t,a_t)}]$ 
through a process of trial and error, expressed as a sequence of states and actions $\tau$.
The main objective of an RL agent can be shown as:

\begin{equation} \label{pi_star}
    \pi^* = \argmax_\pi J(\pi)
\end{equation}


Similar to several related works~\cite{chen2023adapt,rosenstein2005transfer,finn2018probabilistic}, 
this paper focuses on applying policy gradient RL methods~\cite{sutton1999policy},
due to its demonstrated sample efficiency and 
ability to find high-quality solutions in our problem.
Policy gradient methods parameterize policy $\pi$ as
$\pi_\theta(a|s, \theta){=}Pr(a_t{=}a|s_t{=}s, \theta)$
via a neural network;
where $\theta$ is the policy's parameters, 
and $Pr(a_t{=}a|s_t{=}s, \theta)$ denotes the probability of taking action $a$ in state $s$ at time step $t$.
Policy gradient methods optimizes Equation~\ref{pi_star}
by iteratively updating $\theta$. 

A widely used class of policy gradient algorithms for our problem are actor-critic method
\footnote{These methods are considered hybrid, combining value and policy-based methods.}. 
In these methods, actor is a policy function $\pi_\theta(a|s, \theta)$,
and critic is a value function, that estimates $J(\pi_\theta(a|s, \theta))$.
The critic's parameters are updated with respect to 
minimizing the difference between the actual rewards received and its estimates.
By combining these functions, 
policy $\pi_\theta(a|s, \theta)$ is converged towards an optimal policy $\pi^*$,
solving Equation~\ref{pi_star}.

\subsection{GFlowNets and its Variant CFlowNets}
\label{gflownets_background}
Generative flow networks (GFlowNets)~\cite{bengio2021gflownet}, 
are deep learning models for exploratory control tasks. 
In its most basic form, it is a trained stochastic policy or a generative model. 
GFlowNets use two sampling processes, i.e., forward-sampling and backward-sampling
In the forward-sampling policy, the generative model is used to generate distribution and sample candidate object $x$, 
through a sequence of constructive steps (individual actions taken by the policy) with probability proportional to rewards over terminal states. 
Forward sampling policy refers to the method by which the model generates samples from the learned distribution 
i.e., how to transit from one state to the next state,
starting from an initial state $s_0$ until a terminal state $x{\in}\cal{X}$
(with $\cal{X}$ being the space of objects) is reached. 
The candidate objects represent the output of the model's sequential decision-making process.
In backward sampling, 
it starts from a terminal state $x$ with high expected reward,
and iteratively predicts the most probable action that could have led to the current state.
The main components of GFlowNets are as follows~\cite{DBLP:journals/corr/abs-2201-13259}:
  \begin{itemize}
      \item A \emph{forward-sampling policy} that provides the forward action distribution $P_F( - | s )$, where $s\in\mathcal{S}$ is any state that is not terminal. 
      \item A \emph{backward-sampling policy} $P_B( - | s )$ that provides the backward action distribution for backward sampling from any given state $s$.
      \item The initial state flow estimator $Z=F(s_0)=\sum_x \mathcal{R}(x)$, where $s_0$ is the initial state and $\sum_x \mathcal{R}(x)$ is 
      the sum of all the terminal flows at the terminal states.
      \item The edge flow function $F(s{\rightarrow}s')$, which estimates the flow of the edge between two states i.e. the transition between states $s$ to $s'$.
      \item $F(s)$ which is the state flow function that estimates the flow through any particular state.
      \item A self-conditional flow function $F(s|s')$ that estimates flow through $s$ if trajectories only pass through $s'<s$.
      \item A training objective function: Flow matching objective.
  \end{itemize}

\subsubsection{GFlowNets Architecture}
\begin{figure}[!b]
        \centering
        \includegraphics[width=1\linewidth]{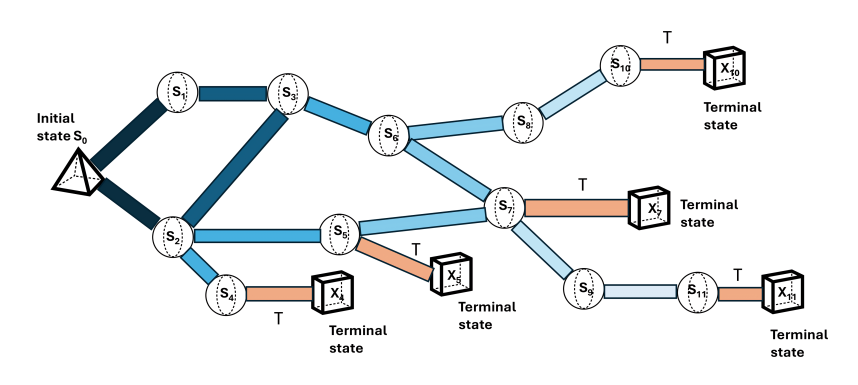} 
        \caption{Flow Network DAG Illustration based on ~\cite{bengio_malkin_jain_2022}.}
        \label{fig:nn_gflownet_animatation}
\end{figure}

GFlowNets can be represented using Directed Acyclic Graphs (DAGs), 
where nodes denote the different states $s_t \in \mathcal{S}$, 
and edges represent actions $a_t \in \mathcal{A}$ (see Figure~\ref{fig:nn_gflownet_animatation}).
a neural network typically samples actions that progressively build the candidate object.
Each executed action given a state $s_t$ in the trajectories transits to a next state $s_{t+1}$ 
with transition probability $F(s_t{\rightarrow}s_{t+1})$. 
When a sample object is constructed, 
it initiates an “exit” action, which leads it to the terminal state, 
after which we can get a reward $\mathcal{R}(x)$.
Then, GFlowNets sample the terminal object with probability proportional to this reward.

The neural network in a GFlowNet
has the capability of outputting a stochastic policy $\pi(a_t | s_t)$. 
the policy results in a forward transition probability $P_F(s_{t+1}|s_t)$. 
In every step, the same neural network is utilized again and it produces a stochastic output $a_t$, 
from which the next state $s_{t+1}=T(s_t,a_t)$ is derived, with $T(s_t,a_t)$ being a transition function 
(similar to $\mathcal{P}(s_{t+1}|s_t, a_t)$) 

GFlowNets formulation is originally designed for discrete tasks, 
in which there are a limited number of state and action pairs.
It is important to note that, most tasks in real environments, 
such as robotic tasks, 
have continuous states and action spaces.
Consequently, in this paper, we choose its variation, 
i.e., Generative Continuous Flow Networks (CFlowNets) 
that is specifically designed to handle continuous domains.

\begin{figure}[htp]
        \centering
        \includegraphics[width=1\linewidth]{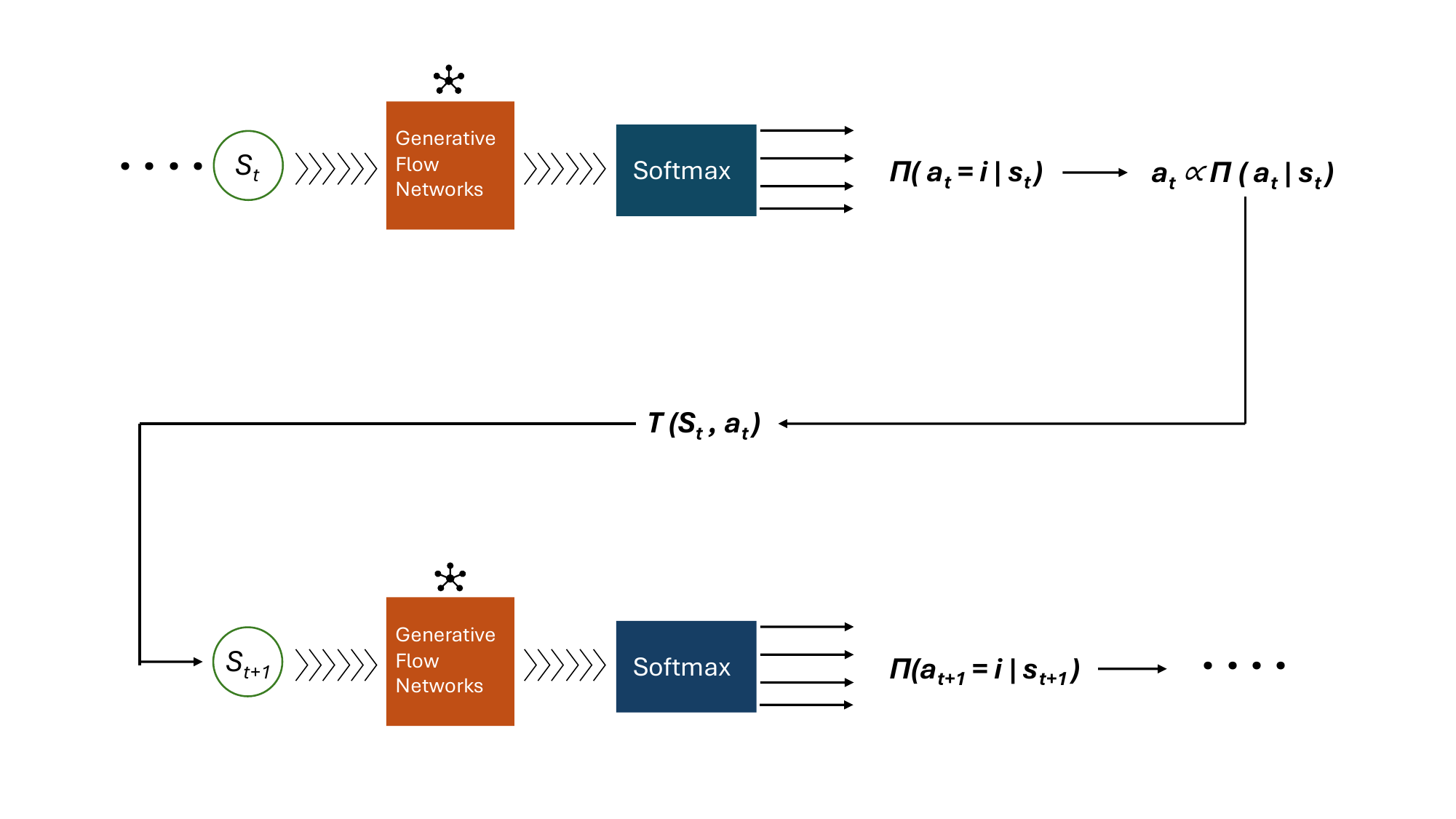}
        \caption{GFlowNets Architecture based on \cite{bengio_malkin_jain_2022}}
        \label{fig:nn_gflownet}
\end{figure}

\subsubsection{CFlowNets Definition}
CFlowNets architecture resembles the architecture of GFlowNets as the following: 
  \begin{itemize}
      \item CFlowNets state space is a continuous state space denoted as $\mathcal{S}$, and the continuous action space is represented with $\mathcal{A}$,
      \item The agent selects an action $a_t\in\mathcal{A}$ to make a transition $s_{t+1}{=}T(s_t, a_t)$, 
      \item The continuous execution of actions results in a sequence of sampled elements of $\mathcal{S}$ which forms an acyclic trajectory $\tau{=}(s_1, s_2, s_3, . . ., s_n)$. 
      \item The acyclic trajectory $\tau$ is as any sampled trajectory starting at $s_0$ and terminating at $s_n$. The initial root state is denoted by $s_0$, and the final state by $s_n$,
      \item Terminating flow is represented as $F(s{\rightarrow}s_n)$ and a transition $s{\rightarrow}s_n$ is defined as the terminating transition.
      \item Continuous flow networks are composed of the tuple $\mathcal{(S, A}, F)$.
  \end{itemize}

\subsubsection{CFlowNets Training Framework}
\label{CFlowNets_training}

We now describe the training framework of CFlowNets.
For any given sample trajectory $\tau{=}(s_1, s_2,..., s_n)$, 
the forward probability will be the product of each individual forward transitions in the sequence:
   \begin{equation}
       \forall \tau = (s_1, s_2,..., s_n), P_F(\tau) := \prod_{t=1}^{n-1}P_F(s_{t+1} | s_t)
   \end{equation}

Similarly, the backward trajectory probability $P_B(\tau)$ can be obtained as the following:
   \begin{equation}
        \forall \tau = (s_1, s_2,..., s_n), P_B(\tau) := \prod_{t=1}^{n-1}P_B(s_t | s_{t+1})
   \end{equation}

Taking into account that the sampled trajectory $\tau$ is acyclic, 
a parent set $\mathcal{P}(s_t)$ is defined 
which contains the set of all the parent states of $s_t$, 
i.e., all the states that can make a transition to the state $s_t$ by selecting an action. 
Similarly, a child set $\mathcal{C}(s_t)$ contains resulting next states after taking an action from the state $s_t$.
To approximate the flow function $F$, 
a non-negative surrogate model of this function ($\hat{F}$) is used,
which accepts a state and action pair as input.
The surrogate flow function $\hat{F}$ corresponds to a flow 
if it meets the following conditions for continuous flow matching:

   \begin{equation}
       \forall s' > s_0, \hat{F}(s') = \int_{s \in \mathbf{P}(s')} \hat{F}(s{\rightarrow}s')ds = \int_{s:T(s,a) = s'} \hat{F}(s,a : s{\rightarrow}s')ds 
   \end{equation}
   \begin{equation}
       \forall s' > s_f, \hat{F}(s') = \int_{s" \in \mathbf{C}(s')} \hat{F}(s'{\rightarrow}s")ds" = \int_{a \in \mathcal{A}} \hat{F}(s',a)da 
   \end{equation}

   If reward environments are sparse, we can train a flow network that meets the flow-matching conditions to obtain the target flow. Using this as a basis, the CFlowNets' continuous loss function can be derived:

   \begin{equation}
\begin{aligned}
    \mathcal{L}(\tau) {=}\sum_{s_t = s_1}^{s_f} \bigg(&\int_{s_{t-1} \in \mathcal{P}(s_t)} F(s_{t-1}{\rightarrow}s_t)ds_{t-1} - R(s_t) \\
    &- \int_{s_{t+1} \in \mathcal{C}(s_t)} F(s_t{\rightarrow}s_{t+1})ds_{t+1} \bigg)^2
\end{aligned}
\end{equation}

\begin{figure}[htp]
        \centering
        \includegraphics[width=\textwidth]{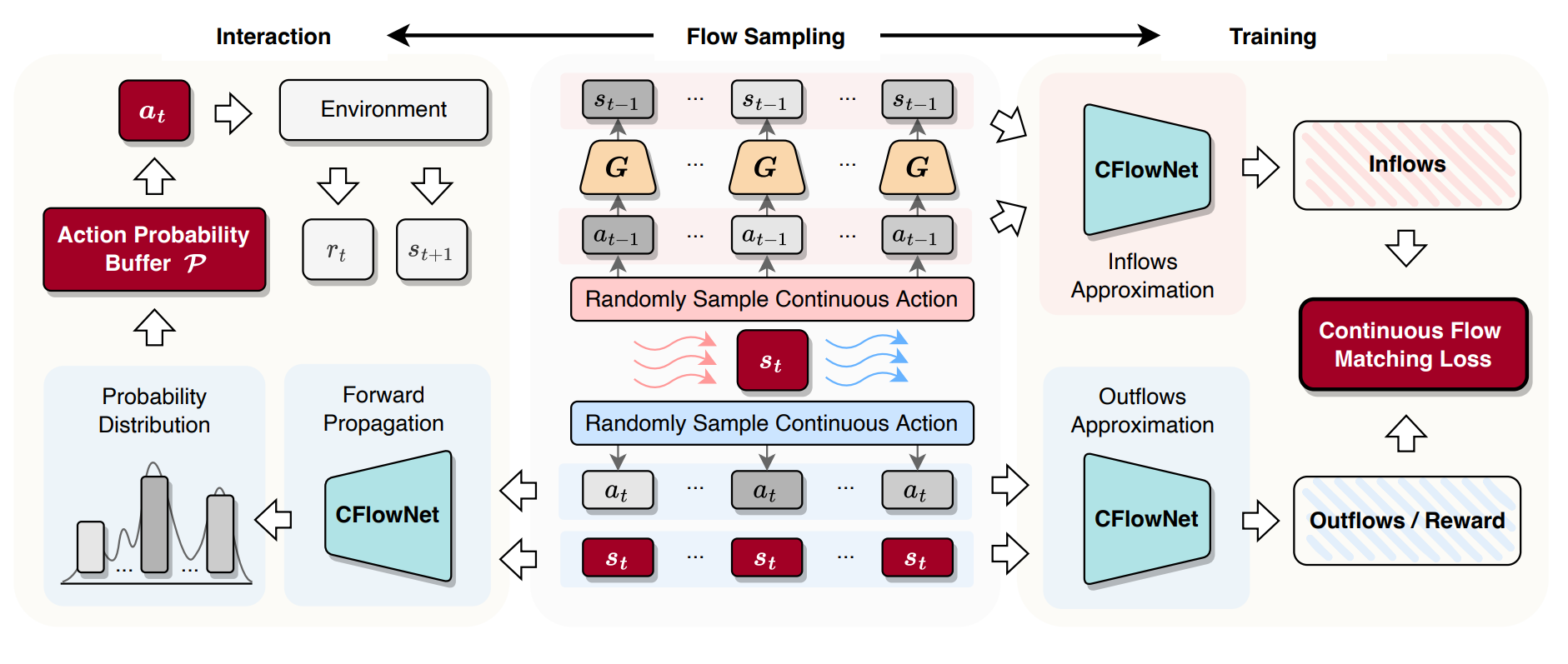}
        \caption{\textcolor{red}{Schamatics of CFlowNets Training Framework \cite{li2023cflownets}. The leftmost part represents the action selection procedure. The middle part is the flow-matching approximation visualization and the rightmost section shows the Continuous Flow-Matching Loss, which is utilized for training.}}
        \label{fig:cflownets training framework}
    \end{figure}

Nevertheless, CFlowNets implementations cannot directly apply this continuous loss function due to the computationally intensive nature of computing integrals over continuous spaces. 
Often these integrals do not have closed-form solutions and require numerical approximation methods.
\textcolor{red}{The overall training framework of CFlowNets can be classified into three parts shown in Figure \ref{fig:cflownets training framework} (additionally see Algorithm~\ref{alg:CflowNets} for details). In the first part, an agent interacts with the environment and how actions are sampled in a continuous action space is discussed. In the second part, flow sampling is done for each state, and in the third part, the CFlowNets are trained using the continuous loss function. The following subsections provide a detailed explanation of these three core components and their roles in the training process}

    \begin{enumerate}
        \item \textbf{Action Selection Procedure:} In the first part, the agent interacts with the environment. 
        The goal is to obtain complete trajectories $\tau$ by iteratively sampling actions from the probability distribution $a_t\sim\pi(a_t|s_t)$, with the help of CFlowNets. 
        The agent first uniformly samples $M$ number of actions from the action space $\mathcal{A}$ at each state $s_t$. 
        Then, it generates an action probability buffer $\mathcal{B}{=}[F(s_t, a_i)]_{i=1}^M $, 
        which approximates action probability distributions. 
        Finally, the agent samples an action from the buffer $\mathcal{B}$ based on the probabilities of all the actions. 
        The actions with higher $F(s_t, a_i)$ will be sampled with a higher probability. 
        After an action is selected, the agent interacts with the environment to update its state. 
        The process is repeated multiple times until the entire trajectory is sampled. 
        The entire process is repeated multiple times to generate sets of trajectories stored in the buffer $\beta$.

        \item \textbf{Flow Matching Approximation:} After obtaining a collection of complete trajectories $\beta$, 
        to satisfy the flow matching condition, 
        it is necessary to ensure for any node $s_t$ that 
        $\int_{a:T(s,a)=s_t} F(s,a) da=\int_{a \in \mathcal{A}} F(s_t,a) da$.
        For calculating the integrals, an approximation method is utilized by randomly and uniformly sampling $K$ actions from the continuous action space $\mathcal{A}$ and calculating the corresponding $F(s_t, a_k)$ values (for $k{\in}\{1,...,K\}$) as the outflows. For outflows, the following approximation is used, where $\mu(A)$ indicates the measure of the continuous action space $\mathcal{A}$:

        \begin{equation} \label{e2}
            \int_{a \in \mathcal{A}} F(s_t,a) da \approx \frac{\mu(\mathcal{A})}{K} \sum_{k = 1}^K F(s_t, a_k)
        \end{equation}

        A deep neural network $G$ (named "retrieval" neural network) is constructed to sample the inflows which is parametrized by $\phi$. The network takes a state-action pair $(s_{t+1}, a_t)$ as the input and outputs the parent state $s_t$. 
        It is trained based on the trajectory buffer $\beta$ with Mean Squared Error (MSE) loss. 
        Thus, the inflows of each state are approximately determined by utilizing the following approximation:

        \begin{equation} \label{e3}
            \int_{a:T(g(s_t,a),a) = s_t} F(g(s_t,a),a) da \approx \frac{\mu(\mathcal{A})}{K} \sum_{k = 1}^K F(G_\phi(s_t,a_k), a_k)
        \end{equation}

        \item \textbf{Continuous Loss Function:} Finally, based on the approximate inflows and outflows in equation \ref{e2} and \ref{e3}, CFlowNets can be trained based on the continuous flow matching loss function:

        \begin{equation} \label{e4}
            \mathcal{L_\theta}(\tau) = \sum_{s_t = s_1}^{s_f} \bigg[\sum_{k = 1}^K F_\theta(G_\phi(s_t,a_k), a_k) - \lambda R(s_t) - \sum_{k = 1}^K F_\theta(s_t, a_k) \bigg]^2
        \end{equation}
        
        $\lambda$ is the scaling factor used to scale the summation of the sampled flows appropriately, considering the size of the continuous action space $\mathcal{A}$.
        
    \end{enumerate}



\begin{algorithm}[H]
    \footnotesize
    \caption{\textcolor{red}{Generative Continuous Flow Networks (CFlowNets)~\cite{li2023cflownets}}}
    \begin{algorithmic}[1]
    \State \textbf{Initialize}: Flow network $\theta$, pretrained retrieval network $G_{\phi}$, and empty buffers $D$ and $P$
    \Repeat
        \State Set $t = 0$, $s = s_0$
        \While{$s \neq \text{terminal}$ and $t < T$}
            \State Uniformly sample $M$ actions $\{a_i\}_{i=1}^M$ from action space $A$
            \State Compute edge flow $F_{\theta}(s_t, a_i)$ for each $a_i \in \{a_i\}_{i=1}^M$ to generate $P$
            \State Sample $a_t \sim P$ and execute $a_t$ in the environment to obtain $r_{t+1}$ and $s_{t+1}$
            \State $t = t + 1$
        \EndWhile
        \State Store episodes $\{(s_t, a_t, r_t, s_{t+1})\}_{t=1}^T$ in replay buffer $D$
        \State [Optional] Fine-tune retrieval network $G_{\phi}$ based on $D$
        \State Sample a random minibatch $B$ of episodes from $D$
        \State Uniformly sample $K$ actions $\{a_k\}_{k=1}^K$ from action space $A$ for each state in $B$
        \State Compute parent states $\{G_{\phi}(s, a_k)\}_{k=1}^K$ for each state in $B$
        \State \textbf{Inflows}: $\log\left(\epsilon + \sum_{k=1}^K \exp\left(F_{\log\theta}(G_{\phi}(s_t, a_k), a_k)\right)\right)$
        \State \textbf{Outflows or reward}: $\log\left(\epsilon + \lambda R(s_t) + \sum_{k=1}^K \exp\left(F_{\log\theta}(s_t, a_k)\right)\right)$
        \State Update flow network $F_{\theta}$ according to the Continuous Loss Function
    \Until{convergence}
    \end{algorithmic}
    \label{alg:CflowNets}
\end{algorithm}

\section{Related Works}
\subsection{Trial-and-Error with select-test-update}
There are several works that proposed trial-and-error approaches for machine fault adaptation
~\cite{chatzilygeroudis2018reset},
leveraging RL algorithms for data-driven adaptability from previously learned policies.
Specifically, Cully et al.\cite{cully2015robots} proposed a trial-and-error approach 
that is inspired by biomimicry, 
where an injured animal can adapt to its injury 
through a process of trial and error to figure out optimal movement. 
To guide the trial-and-error algorithm, 
a pre-computed behavior-performance map was developed that contained 13,000 different robotic gaits. 
Using this map, a robot experiencing any physical malfunction 
uses a simulated behavior with the highest estimated performance value. 
After executing the action in the environment, 
the selected behavior gets a new performance rating for that particular task. 
This process continues until the robot identifies a behavior 
with a performance estimate that exceeds 90\% of the best performance predicted 
for any behavior in the behavior-performance map \cite{cully2015robots}. The reference to the 90\% performance threshold in the paper is a user-defined stopping criterion rather than a rigorously justified value. The experimentation was done on a hexapod robot with six different conditions. 
Using the proposed algorithm with a behavior-performance map, 
the robot was able to adapt to failure by learning compensatory behavior within a reasonably fast time. For instance, for a hexapod robot, the algorithm typically required between 3 to 10 trials, with the adaptation time averaging around 66 seconds for the four damage scenarios tested.


Our study with CFlowNets for fault adaptation introduces a generative model that has the capability of continually learning and adjusting its policy based on new experiences gathered from the environment, which is crucial to incorporating adaptability in dynamic and unpredictable environments. Although the trial-and-error method can be considered adaptable to a certain extent, its performance is heavily dependent on initially pre-computed behavior-performance data which might not convert all possible scenarios and changes in the environment. Additionally, the trial-and-error approach utilizes a behavior-performance map of 13,000 robotic gaits that is computed and tailored to a specific hexapod robot and lacks generalization over different robotic simulations. Our research, on the other hand, involves exploring a framework that can be generalized across different environments and robotic tasks because of its approach to learning distributions over trajectories, thus introducing a higher degree of scalability and generalizability.

\subsection{Adaptation using Meta-RL}
Building on standard RL algorithms, several meta-reinforcement learning were presented
~\cite{ahmed2020complementary,luo2021mesa,nagabandi2019learning},
wherein given a distribution of tasks, 
the objective is to learn a policy, with as little data as possible, 
that adapts to unseen (but similar) tasks
using the task distribution.
Nagabandi et al.~\cite{nagabandi2019learning} proposed using model-based meta-reinforcement learning
for fault adaptation tasks.
In their meta RL framework, two adaptive learners were utilized for the algorithm. 
An initial set of parameters for a generalized dynamics model is learned 
using model-agnostic meta-learning (MAML)~\cite{finn2017model}.  
This algorithm uses a gradient-based adaptive learner (GrBAL) 
whose dynamics model is represented using a neural network. 
The second learner is a recurrence-based adaptive learner (ReBAL) 
that uses a recursive neural network to represent the dynamics model. 
To demonstrate the sample efficiency of both GrBAL and ReBAL, 
Clavera et al. reported the average return in differing test environments 
based on the amount of data used in meta-training. 
The results were compared with two model-free methods,
i.e., Trust Region Policy Optimization (TRPO) and MAML-RL. 
Their proposed GrBAL and ReBAL outperform the baselines 
in terms of sample efficiency 
even though the baseline methods were trained with 1000 times more data. 
Additionally, they conducted a comparative analysis on several continuous control tasks 
including simulated robots 
(OpenAI Gym’s Ant and HalfCheetah) 
and real-life robots (millirobot), 
and the results showed faster adaptation to changing environments.


Similar to this study on adaptation using RL and Meta-RL, our experiments were also conducted on the OpenAI Gym environment, which is a standard and widely utilized framework to evaluate algorithms on continuous tasks. Additionally, similar performance metrics were adopted from this study as we conducted our comparative analysis between CFlowNets and standard RL in terms of learning efficiency, adaptation speed, and average reward performance metric. However, Meta RL often requires a distribution of tasks for training to adapt effectively to new but similar tasks. In contrast, CFlowNets can generalize across different environments and robotic tasks more effectively due to its sampling strategy, which allows it to explore a wider range of possible solutions.

\subsection{CFlowNets for Continuous Control Tasks}
Li et al.~\cite{li2023cflownets} proposed a CFlowNet framework for different continuous control tasks, 
i.e., Point-Robot-Sparse, Reacher-Goal-Sparse, and Swimmer-Sparse.
The rewards associated with these three tasks were sparse.
They compared CFlowNets with state-of-the-art RL algorithms,
i.e., deep deterministic policy gradient (DDPG),
Twin Delayed DDPG (TD3),
Proximal policy optimization (PPO),
Soft Actor-Critic (SAC),
in terms of the reward distribution, average rewards, and the number of valid-distinctive trajectories generated.
Interestingly, their findings demonstrated that 
CFlowNets were highly effective at fitting the true reward distribution,
where the RL algorithms struggled.
To study the exploration behavior of CFlowNets and the RL algorithms,
$10000$ valid-distinctive trajectories\footnote{if two trajectories have high returns, but their MSE is small, then instead of two trajectories, only one is counted} were generated using these methods.
The results showed that DDPG, PPO, and TD3 have little to no exploration ability 
since they only generated one valid-distinctive trajectory. 
SAC demonstrated good exploration behavior at first 
but then the performance decreased as the training progressed. 
On the other hand, CFlowNets showed remarkable exploration capabilities, 
generating thousands of unique valid-distinctive trajectories that far exceeded any other RL algorithm. 
Furthermore, for Point-Robot-Sparse and Reacher-Goal-Sparse tasks, 
CFlowNets achieves a better average return in fewer timesteps than the RL algorithms. 
However, it did not perform well in the Swimmer-Sparse task. 

It is important to note that,
the reason behind the superior performance of CFlowNets, as explained by the authors, 
is that both the Point-Robot-Sparse and Reacher-Goal-Sparse have evenly distributed rewards 
requiring more exploration.
The exploration strategy of CFlowNets aligns well with
exploration-based tasks, such as machine fault adaptation,
which often involves subtle changes in the machine's state, e.g., actuator faults.
The ability to actively explore promising regions,
CFlowNets could potentially recover from faults quickly and find (near) optimal solutions.
Given the strong performance of CFlowNets reported in this research,
it is worth investigating how they perform during machine fault adaptation,
where sample efficiency and quality of solutions are crucial.

While the study by Li et al.~\cite{li2023cflownets} demonstrated the efficacy of Continuous Flow Networks (CFlowNets) for various continuous control tasks, our work extends this foundation in several significant ways. Li et al. focused on general continuous control tasks with sparse rewards. In contrast, the primary focus of this research was in fault adaptation in robotic systems. We introduce and evaluate CFlowNets in four custom gym environments with simulated faults that mimics real world machine faults. This is an important aspect of real-world robotic applications where reliability and fault tolerance are of utmost importance. Additionally, We investigate different knowledge transfer options by transferring knowledge from a pre-fault environment to a post-fault environment. This phase of our experimentation demonstrates how pre-learned policies and stored experiences can accelerate adaptation to OOD situations, and unseen fault-induced conditions, providing insights into the practical deployment of CFlowNets in dynamic scenarios.

\section{Methodology}
This section details our methodology, 
providing an in-depth description of the experimental setup 
and the robotic environments employed. 
Furthermore, we discuss each of the four faults 
that were introduced in the simulated robotic environments 
and how these faults are applied to create four custom gym environments. 
Then, we describe our implementation of CFlowNets 
and compare it with several state-of-the-art RL algorithms.
To ensure reproducibility,
We provide detailed descriptions of both our experimental designs and evaluation metrics used,
and make our code publicly available at \url{https://github.com/zahinsufiyan/Flow-Networks-Fault-Adaptation}.

\subsection{Experimental Setup}
We present our experimental setup in Figure~\ref{fig:method}. 
For our comparative study, 
our experimental setup consists of three main stages. 
In Stage 1, an agent learns a robotic task in a normal environment for 10 million timesteps to ensure convergence. 
In this paper, we explore using a CFlowNet as our agent in this task.
For comparison, we use state-of-the-art RL algorithms in this field (DDPG, TD3, PPO, and SAC) as our baselines.
At the end of Stage 1, we store the learned knowledge, i.e., policy parameters and memory buffer/experience buffer.
In Stage 2, we introduce four different faults that often robotic systems face in real-world applications.
In Stage 3, the agent attempts to learn the same robotic task in Stage 1, 
but in a faulty environment 
(one of the faults in Stage 2).
In this stage, we investigate if the learned knowledge from Stage 1
contains useful information resulting in a faster fault adaptation 
and potentially finding a better solution in faulty environments.
Therefore, the policies learned in Stage 1,
are directly transferred to Stage 3 
(thus, the early time steps in this stage reflect the performance of the policies learned in Stage 1). 
It is important to note that at the start of this stage,
we assume the system has the capability to detect and diagnose the faults,
thus the focus of this research is on fault adaptation.

\subsubsection{\textcolor{red}{Dataset and Environment}}
\begin{figure}[tb]
\centering
    \includegraphics[width=1\textwidth]{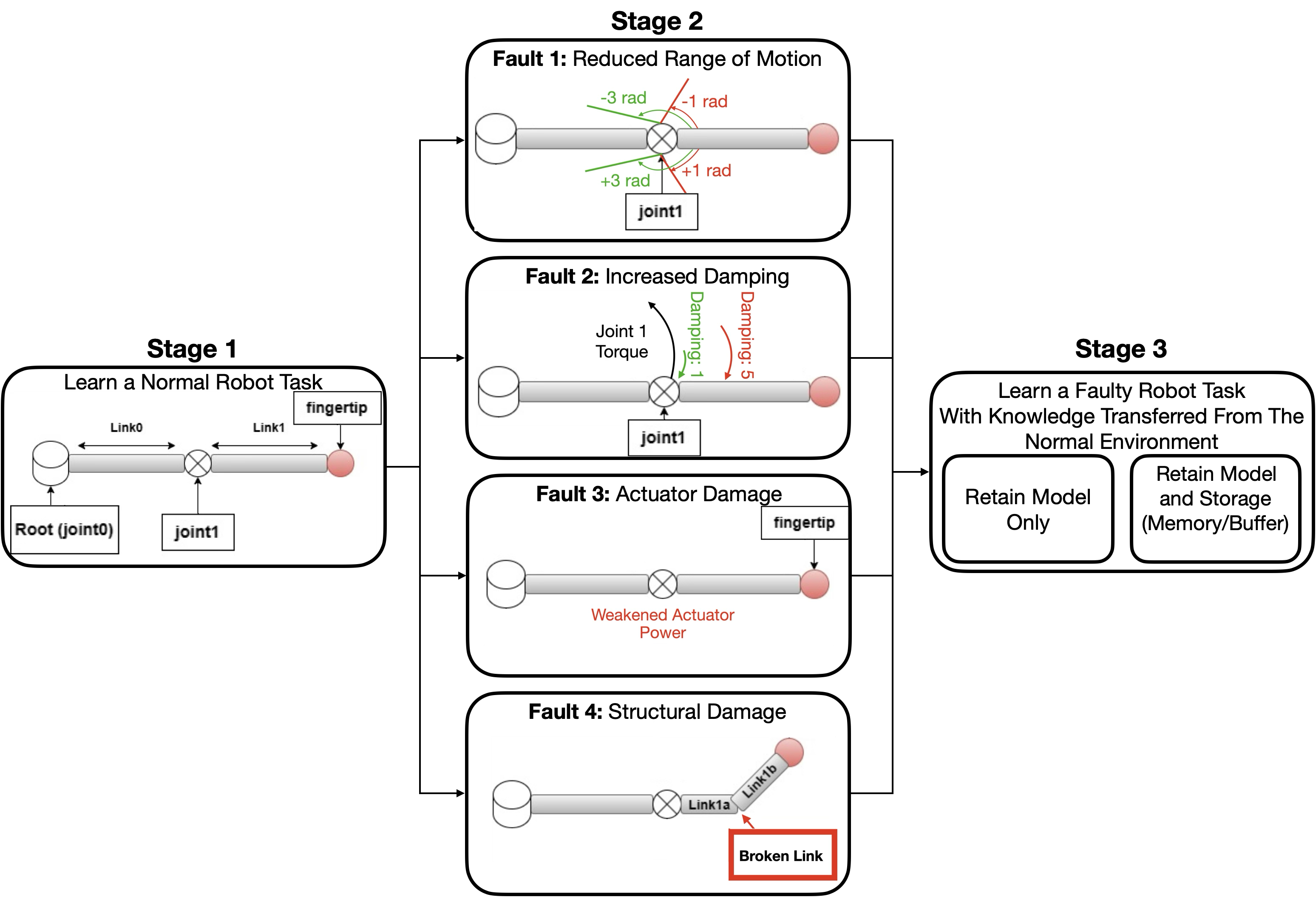}
        \caption{\textcolor{red}{An Overview of Our Experimental Setup.}}
        \label{fig:method}
\end{figure}

Similar to several related work
~\cite{fournier2018accuracy,li2023cflownets,parisotto2019concurrent}, 
we opted for the Reacher-v2 robot arm within MuJoCo (Multi-Joint Dynamics with Contact). 
In this study, the dataset was generated from this Reacher-v2 simulated robotic environment, implemented in the MuJoCo simulator.
Mujoco is a physics engine used for research and development in 
robotics, biomechanics, graphics and animation, machine learning, and other areas \cite{1606.01540}.
The reason for choosing this robot arm is that
it is a well-defined robotic system,
facilitating a manageable introduction of faults,
and evaluation of how different algorithms can handle the faults.
The Reacher-v2 environment is simulated in a 2D plane and 
consists of a two-joint robotic arm with an end effector/fingertip at the end of the arm. 
The two joints are named joint0 and joint1 in Figure~\ref{fig:method}. 
These joints are capable of a wide range of motion. 
Joint0 fixes the first part of the arm (link0) 
with the point of fixture also known as the root, and 
joint1 connects the second part of the arm (link1) to link0. 
The robot arm uses actuators/motors to control these joints, 
which help provide torque so it can maneuver around the 2D plane.
The objective of the task is to maneuver the end effector or fingertip 
to reach a specified target location which is defined in the environment. 
With each episode, the coordinates of the target change to various locations within the environment.  

\textcolor{red}{
We modified this environment to introduce four fault scenarios that represent real-world mechanical malfunctions. 
To do so, the first step is to modify the XML file of the Reacher-v2 environment,
used by the MuJoCo physics engine to create a simulated environment. 
This file is a configuration file containing the physical parameters and the structure of the Reacher robot and its environment. 
Users can simulate the robot's movement and interaction in a virtual environment. 
The dataset exhibits diverse state-action distributions due to varying mechanical conditions which are introduced through the faulty environments, reflecting the robot's adaptive strategies under different constraints. 
These conditions result in distributions that add variability which in turn enriches the dataset. 
Specifically, the recorded trajectories for each fault differ from those of other faults. 
For instance, the same set of actions taken in a given state 
is likely to lead the agent to different sets of subsequent states.
As a result, the dataset distribution reflects dynamic conditions that challenge the model to generalize across a wide range of adaptive behaviors and provide a robust basis for evaluation.}

\textcolor{red}{
The dataset consists of recorded trajectories from normal and faulty environments. 
Each trajectory includes state, action, and reward data across a time horizon 
where the agent learns to reach the target. 
Data was gathered by running simulations over multiple episodes for each fault scenario,
accumulating over 10 million timesteps across all experiments. 
This scale captures variations in the robot's movement under each fault condition, 
providing a rich dataset for evaluating adaptive performance.}

\textcolor{red}{
Several preprocessing steps were taken to maintain consistency across all training conditions. 
For instance, state variables 
representing joint angles, velocities, and fingertip-to-target vectors 
were normalized to ensure uniformity across fault conditions. 
The reason for this normalization was to ensure 
the model would effectively generalize since we introduced fault scenarios of varying scales. Additionally, actions were clipped to ensure that 
control limits were maintained within each type of fault configuration 
so that the values remained within the physical constraint. 
For example, while introducing actuator damage 
(Fault 3 in Figure~\ref{fig:method}) 
by reducing motor gear ratios, 
the environment required stricter control limits to simulate degraded motors with accuracy. Furthermore, the position of the target was randomized at the start of each episode within some range. 
The randomization in position is controlled so as not to bias towards any particular position of the target and to keep the target within reach, 
even under very restrictive fault conditions.
}



\subsection{Stage 1: Learn a Normal Robot Task}
The primary objective is to investigate 
whether CFlowNets can adapt when a fault occurs.
To explore this, we first compare the performance of CFlowNets and our baselines
on a normal robot task, wherein no faults are present.
At the end of this stage, we store the knowledge that these algorithms learn from the environment, 
i.e., models parameters and the experienced trajectories.
We transfer the knowledge to Stage 3, enabling us to study the adaptation capabilities of the algorithm.


\subsection{Stage 2: Fault Injection}
We design four different faults in this stage. 
Each of these faults represents and mimics a real-world malfunction that 
are encountered by robot arms in a practical environment. 
We created four different custom gym environments representing these four faults. 
Each of these four custom gym environments contains a constant fault type,
i.e., values of the modified attributes remain constant over time. We assume that the faults introduced in the robotic environments are sudden and not progressive, which means the faults occur abruptly without prior indication of gradual faults.
To determine the modified values of the attributes, 
an attribute search (similar to hyperparameter search) 
was conducted considering the severity of the faults 
in such a way that the fault introduces more complex dynamics in the new environment, on one hand,
but still allows meaningful learning and adaptation.
This iterative testing approach is similar to a hyperparameter search where we test different settings of attributes and observe the learning trend for a limited number of timesteps. The severity of the faults was chosen in such a way that the fault introduces more complex dynamics in the new environment. However, the new dynamics of the environment were not changed to such an extent that completely derailed the learning process of the algorithms. In other words, the particular attribute values chosen for simulating the four fault environments were intended to significantly impact the Reacher-v2’s performance but still allow for meaningful learning and adaptation. We started with random initial values(slightly deviated from the original values) that were later modified based on preliminary runs to ensure that the fault produces a substantial amount of challenge for the robotic arm but does not render the task unsolvable. By performing this iterative attribute search and fine-tuning the attribute values we were able to create four custom gym environments to test each of the algorithms' adaptive performance. In the following parts, we discuss each of these faults and how they were simulated in the environment.

\paragraph{\textbf{Fault 1: Reduced Range of Motion (ROM)}}
Robotic arms can experience a reduced range of motion or 
joint angular displacement due to a number of factors, 
including gear wear and tear, 
mechanical restrictions, 
or malfunctioning software \cite{motioncontroltips2023, universalrobots2022, universalrobots2022overview}. 
This type of fault impacts the precision and flexibility of a robot in maneuvering its joints. 
In the context of Reacher-v2, 
a reduction in its ROM will lead to poor performance 
as the arm will not be able to reach its distant targets 
which may require a wide range of motion within the given limited timestep. 
To simulate this fault, 
we change the range of \textit{joint1} from $[-3.0,~3.0]$ radians to $[-1.0,~1.0]$ radians 
(see Figure~\ref{fig:method}), 
which means the joint is not able to rotate to its full extents.


\paragraph{\textbf{Fault 2: Increased Damping}}
Damping refers to the decrease in oscillation amplitude. 
When there is an increase in damping in the robot’s joint 
it indicates that there is an abnormal resistive force 
which is impairing the oscillation of the robot’s links. 
In a practical environment, 
this type of malfunction may be caused by outside contaminants 
affecting the joints or a mechanical problem that increases friction within the joints \cite{hindawi2023friction, asme2023static, cambridge2023modeling}. 
To simulate this fault, we increase the damping of joint1 from $1$ to $5$ 
(see Figure~\ref{fig:method}).
    

\paragraph{\textbf{Fault 3: Actuator Damage}}
Robots have actuators/motors within their joints that 
provide torque for maneuvering the joints. 
Actuator damage can be caused by gear damage, overheating, electrical faults, etc \cite{IndustrialAutomation2023servo, KEBAmerica2023}. 
Due to this type of fault, 
the Reacher-v2 robotic arm may experience incorrect movements, 
reduced output force at the joints, 
and an inability to maintain position. 
We simulate the actuator damage fault by decreasing the actuator power from $200$ to $100$,
mimicking a weakened motor.


\paragraph{\textbf{Fault 4: Structural Damage}}
One of the most common types of fault in real-world application of robotics 
is the structural damage of robotic manipulators. 
Structural damage can occur for various reasons 
such as external impacts, corrosive elements, manufacturing defects, 
or simply because of the deterioration of the materials.
To introduce a structural damage fault in Reacher-v2, 
we simply bend link1 $45$ degrees (see Figure~\ref{fig:method}). 

\subsection{Stage 3: Learn a Faulty Robot Task}
\textcolor{red}{Stage 3 is an ablation study to examine how} CFlowNets and our baseline methods adapt to various faults 
while retaining their model parameters \textcolor{red}{(policies learned in the normal environment in Stage 1)}. 
We compare their performance in terms of adaptation speed, and sample efficiency.
The model parameters hold the knowledge learned in the pre-fault environment (Stage 1),
ensuring the model is not required to learn the task from scratch.

Additionally, we transfer the storage (replay buffer/memory) contents 
that were saved while training in Stage 1 to Stage 3 
which can potentially accelerate learning by leveraging prior experiences. 
The storage contents include the state, action, reward, and next states of each experience, 
and represents the agent’s interaction in the pre-fault environment (Stage 1).
Thus, when deployed in the fault environment as a starting point,
the storage contains experiences from the original pre-fault environment and 
also new experiences that the agent gathers in the new post-fault environment.
By replaying experiences, instead
of learning from scratch, the replay buffer enables the agent to exploit good
policies from the start without random exploration.
\textcolor{red}{Note that in similar studies, such as~\cite{schoepp2024enhancing}, training the agent from scratch or retaining only the collected experiences perform worse than transferring the policy with or without experiences. Therefore, we did not consider these configurations in our ablation study.}


\subsection{Algorithms Implementation and Experiments Details}
We now provide a brief overview of the algorithmic implementation of CFlowNets and 
the four RL algorithms (DDPG, TD3, PPO, and SAC) as our baselines.
We run all of the algorithms for 10 million timesteps to determine where each of them converges to their asymptotic
performance. There were 50 timesteps in each episode of the task. 
The policy evaluation is performed for all the algorithms 
in each 5,000 timestep interval, 
meaning that at each interval, 
we freeze the learned policy and evaluate it for 10 episodes.
The rewards received in this evaluation period are averaged and 
recorded for further analysis. 
We run each experiment 10 times, 
and the performance variability across these 10 runs 
is shown in the shaded regions which correspond to a 95\% confidence interval.
The experiments are performed on Ubuntu 20.04.6 LTS server running Linux, equipped with NVIDIA RTX A6000 GPUs.

\paragraph{\textbf{CFlowNets}}

According to Section~\ref{CFlowNets_training},
the training framework of CFlowNets has three parts.
A retrieval network is constructed in order to make predictions about the parent nodes of each state in a sampled trajectory, which is elaborated in Equation \ref{e3}. The network is made of a feed-forward neural network. This NN consists of four layers. The input layer concatenates the state and action vectors in the first input layer. In each hidden layer, we have 256 neurons with ReLU activation functions applied. Tuples of experience are stored and managed in this experience buffer. The Adam optimizer with a learning rate of 0.003 is used for network optimization.

Subsequently, the Retrieval network is employed to enhance the precision of locating parent states during flow matching in the implementation that involves the training of the flow network. The architecture of the flow network comprises a feedforward neural network with three layers. The concatenation vector of state and action is fed from the input layer through two hidden layers. 
There are 256 neurons in each hidden layer with the ReLU activation function applied. 
As the final layer, the output layer is composed of a single neuron that outputs the calculated edge flow. 
Using Softplus activation functions in the output layer ensures that edge flow values are non-negative since flows cannot be negative. 
For the flow network, the Adam optimizer with a learning rate of 0.003 is used for network optimization.

CFlowNets implementation involves several configurable hyperparameters, including action probability buffer size, number of sample flows, policy noise, noise clip, etc. We conducted a hyperparameter search for the CFlowNets algorithm to figure out the best combination of hyperparameters that leads to optimal performance. The search began with the published CFlowNets hyperparameter settings. Then we compared different hyperparameter settings by recording the average return in the last 200 evaluation for a single run. This process was conducted for a total of 10 runs to account for the performance variability. The rewards from these 200 evaluation points across 10 runs were then averaged again to get a single performance metric for each set of hyperparameter settings. This final average return is then used to compare the performance under different set of hyperparameters. A higher average return signifies a better hyperparameter setting. 

Tensorboard was also used for real-time training monitoring, which aided in fine-tuning hyperparameters. It is worth mentioning that all the hyperparameter tuning was done for the Reacher-v2 normal environment and not for the four modified fault environments. This decision was made based on the assumption that in practical scenarios we are optimizing the model only for the normal environment and not for the fault environments. The list of selected hyperparameters is shown in Table \ref{tab:hyper-cflownets}.

\begin{table}[htbp]
\centering
\caption{Best Performing Hyperparameters in the CFlowNets training.}
\label{tab:hyper-cflownets}
    \begin{tabular}{| c | c |}
        \hline
        CFlowNets & Reacher-v2 \\
        \hline
        Total Timesteps & 10,000,000 \\
        Max Episode Length & 50 \\
        Eval Frequency & 5,000 \\
        Learning Rate & 0.003 \\
        Batchsize & 256 \\
        Retrieval Network Hidden Layers & [256, 256, 256] \\
        Flow Network Hidden Layers & [256, 256] \\
        Number of Sample Flows & 100 \\
       Action Probability Buffer Size & 10,000 \\
        Replay Buffer Size & 100,000 \\
        $\epsilon$ & 1.0 \\
        Optimizer & Adam \\
        \hline
    \end{tabular}
\end{table}


\paragraph{\textbf{RL Algorithms (DDPG, TD3, SAC, PPO)}} 
We compare CFlowNets' performance with state-of-the-art RL algorithms: DDPG, TD3, SAC, PPO. 
We borrow the implementation of both DDPG and TD3 from~\cite{fujimoto2018addressing}. 
For our implementation, we used the published hyperparameters for DDPG and TD3 in \cite{lillicrap2015continuous} and \cite{fujimoto2018addressing}, respectively, 
with minor modifications to the coefficient for soft update ($\tau$) and policy noise. ($\tau$) affects how rapidly the target networks are updated towards the main networks, which can impact the stability and convergence rate of the learning process. Policy noise, on the other hand, adds a level of exploration to the policy actions.

For Soft-actor-critic (SAC), we use the implementation available at~\cite{pytorch_sac}, 
which is derived from the SAC's original paper~\cite{haarnoja2018soft}.
The actor’s network is a three-layered network with $256$ neurons in each of the hidden layers. 
The learning rate of both actor and critic networks is $0.0001$. 
For the entropy term, the initial temperature parameter $\alpha$ is set to $0.1$, 
and learnable temperature is enabled, 
allowing the model to optimize temperature during training.

For PPO, we use the Stable Baselines3 \cite{stable-baselines3} implementation with a few minor modifications. 
In the stable baselines3 version, no explicit evaluation frequency parameter was included. 
We added this feature in our implementation 
to enable stopping the training after a specified number of iterations, 
evaluate the current policy for a specified number of steps, 
and record the performance of the current policy. 
The code implements the clipped surrogate objective. 
The "clip" parameter controls the extent to which the policy can be updated in a single step, 
providing a form of regularization and ensuring stability in training. 
We apply a “clip\_vf” parameter 
which enables the clipping mechanism to the value function. This parameter helps in controlling the magnitude of updates, preventing extreme changes to the value estimates and ensures learning stability.
Additionally, we added the feature of linear learning decay options and 
used the generalized advantage estimator (GAE)~\cite{mnih2016asynchronous}, 
resulting in more stable policy updates. 

\section{Results}
This section presents our experimental results 
for our comparative analysis of CFlowNets and the RL algorithms in the field of robotics and fault adaptation.
Our results are presented in five distinct segments: 
1) adaptation performance,
2) adaptation speed and sample efficiency,
3) execution time,
4) GPU memory usage, and
5) CFlowNets’s Model and Storage Transfer Analysis.



As mentioned, we run each experiment for 10 million time steps; 
but for more clarity and readability of performance during the early stages of learning, 
the plots are truncated to keep the first 1 million time steps.

\subsection{Adaptation Performance}
In the first segment of our comparative analysis, 
we evaluate the adaptation performance and sample efficiency of CFlowNets compared to RL algorithms 
after the introduction of each fault from Stage 2. 
For more effective analysis, 
we categorize the faults into pairs: 
Fault 1 and Fault 2 as Motion Impairment Faults, and 
Fault 3 and Fault 4 as Structural and Mechanical Faults. 
In this section, the model parameters learned in Stage 1 is deployed in Stage 3 
to evaluate each algorithm's adaptability while experiencing a fault.

\begin{figure}[htbp]
    \centering
    \begin{subfigure}[b]{0.8\textwidth} 
        \centering
        \includegraphics[width=\textwidth]{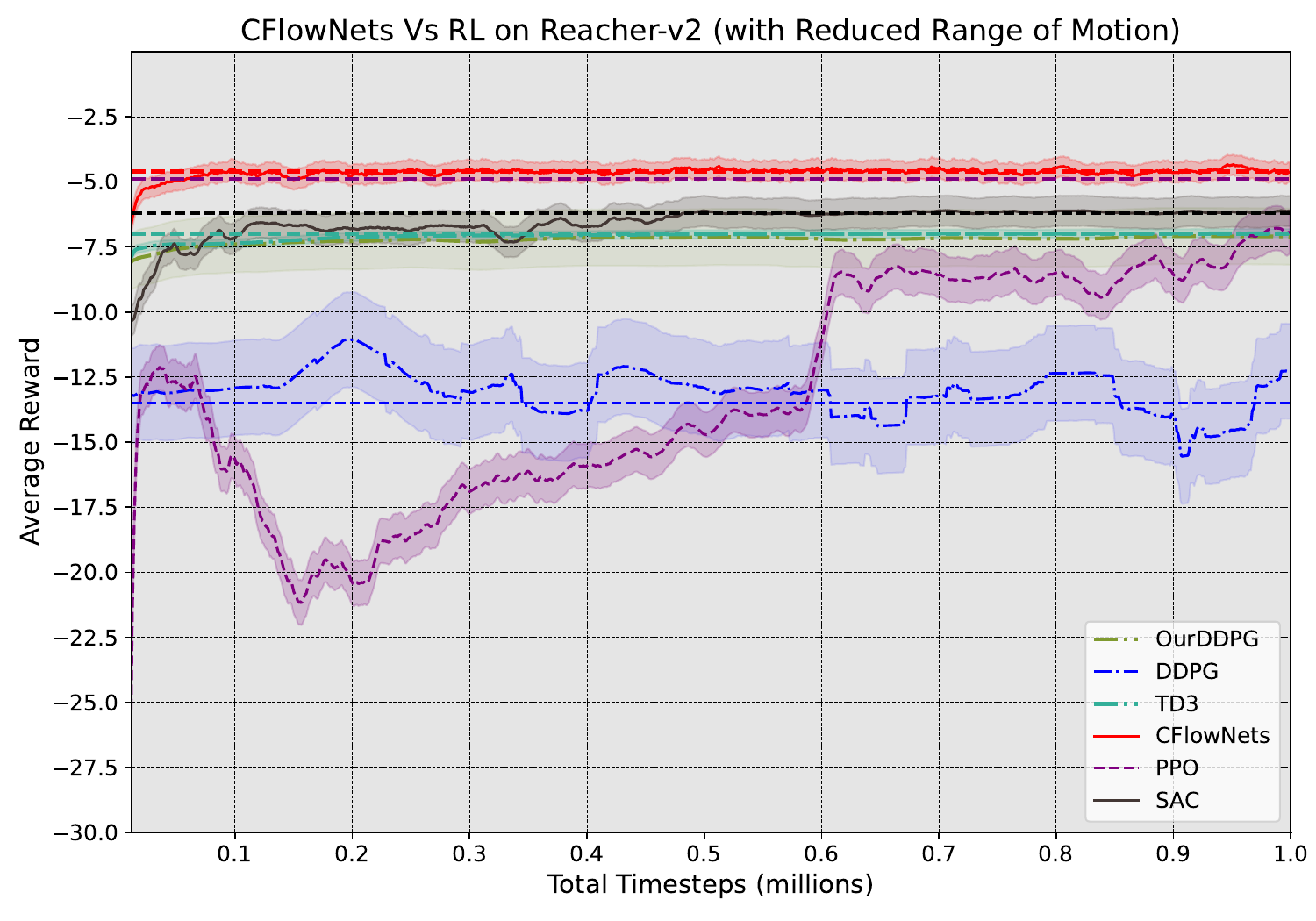}
        \caption{Reduced Range of Motion (Fault 1)}
        \label{fig:a_ROM}
    \end{subfigure}
    
    \vspace{0.5cm} 
    
    \begin{subfigure}[b]{0.8\textwidth} 
        \centering
        \includegraphics[width=\textwidth]{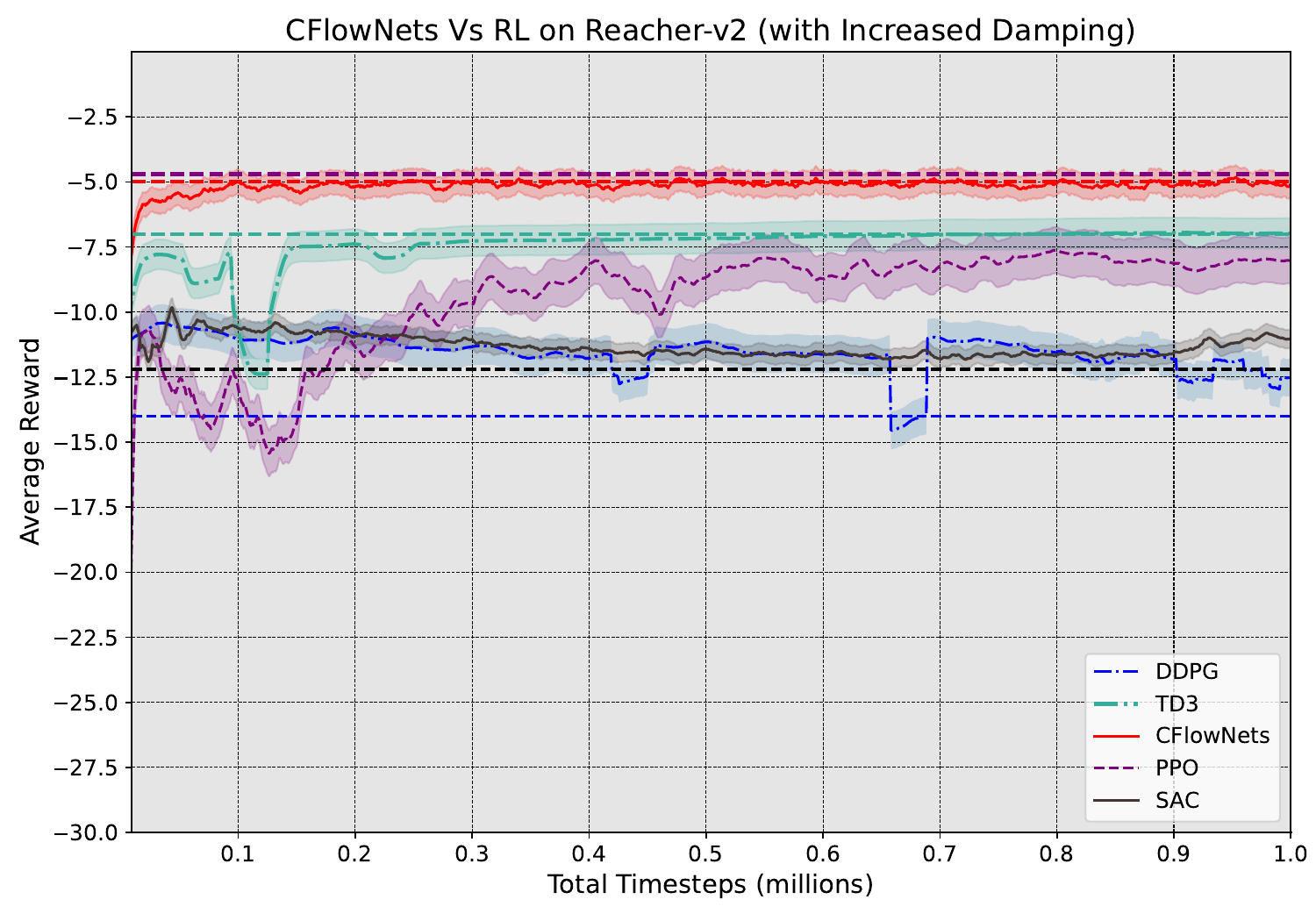}
        \caption{Increased Damping (Fault 2)}
        \label{fig:b_Damping}
    \end{subfigure}
    
    \caption{The early performance in Motion Impairment Fault environments is depicted through learning curves for all five algorithms. The dashed line represents the asymptotic performance.}
    \label{fig:ab}
\end{figure}

\paragraph{\textbf{Motion Impairment Faults (Fault 1 and Fault 2)}} 
Figure~\ref{fig:ab} depicts the average return 
with respect to the number of real experiences collected (i.e., time steps) 
for the environment with Fault 1 and Fault 2 (reduced range of motion and increased damping respectively).
The highest asymptotic performance for each method is shown with a dashed line,
with each line colored the same as its corresponding method.
According to Figure~\ref{fig:ab}, 
we can see that CFlowNets maintains a relatively consistent performance for both fault environments. 
The narrow shaded area corroborates the stability and minimal fluctuation of the algorithm over the ten trials. The narrow shaded region represents CFlowNets' minimal variability, a crucial aspect in fault-tolerant applications where stable performance is required. Additionally, it surpassed all the other RL algorithms 
in terms of the highest asymptotic performance (average reward), 
accumulating the highest average reward compared to all the RL algorithms for Fault 1.
This observation indicates that CFlowNets architecture has the inherent capability 
to understand and adapt in adversarial conditions while dealing with constrained action space. 
For Fault 2, Figure \ref{fig:b_Damping} indicates that while CFlowNets did not top the charts, 
its performance remained robust, closely mirroring the PPO’s asymptotic performance. 
Based on sample efficiency, CFlowNets provided excellent results because, for both fault environments, CFlowNets required the least amount of real experience to gain a near-asymptotic performance. In this scenario, the inherent structure of CFlowNets appears well-suited to handling friction-induced damping effects, likely due to its efficient trajectory sampling strategy. This enables CFlowNets to prioritize high-reward actions while preserving stability, which is evident from its relatively consistent average reward across trials.

Among the baseline algorithms, PPO showed the greatest discrepancy 
between its asymptotic performance and performance under Fault 1 and Fault 2 conditions. 
As Figure~\ref{fig:ab} shows, PPO achieved a comparable asymptotic performance, while
in both Fault 1 and Fault 2, 
it encounters an initial dip in performance and undergoes a steeper learning curve. 
However, as the time steps progressed, 
the PPO algorithm was able to stabilize its learning curve 
after collecting a considerable number of real experiences. 
It maintained a high average reward at the end of the learning period of 10 million time steps, 
gained convergence closer to CFlowNets’s learning curve for Fault 1. 
For Fault 2, PPO’s eventual convergence to the highest asymptotic performance among all algorithms suggests that, despite requiring more samples initially, PPO adapts well to damping-related constraints. 
Notice that across the 10 experimental runs, 
PPO has a wider shaded region, 
indicating a variance in its performance in response to environmental changes, which may be less desirable in applications requiring consistent performance.

TD3 and SAC exhibited similar performance in the environment with the reduced range of motion. 
Although SAC outperformed TD3 in terms of average return with a higher learning curve, 
TD3 displayed better sample efficiency by achieving asymptotic performance faster. 
However, the narrative changed when both TD3 and SAC algorithm was run in the Fault 2 environment. 
TD3 outperformed SAC both in terms of a higher learning curve and better sample efficiency. 
Although TD3 faced some performance setbacks in its initial learning phase, 
it quickly stabilized within 0.5 million time steps. Conversely, SAC experienced a decline in its learning curve and necessitated a significantly greater number of timesteps to stabilize and converge.

DDPG, the predecessor of TD3, 
exhibited the worst performance. 
DDPG showed volatile and fluctuating performance since the beginning of its learning phase and 
showed numerous dips in its performance throughout the 10 million time steps. 
This downward and unstable trend suggests poor policy modification for adapting to these faults, 
limiting its practical application in such tasks.

\begin{figure}[htbp]
    \centering
    \begin{subfigure}[b]{0.8\textwidth} 
        \centering
        \includegraphics[width=\textwidth]{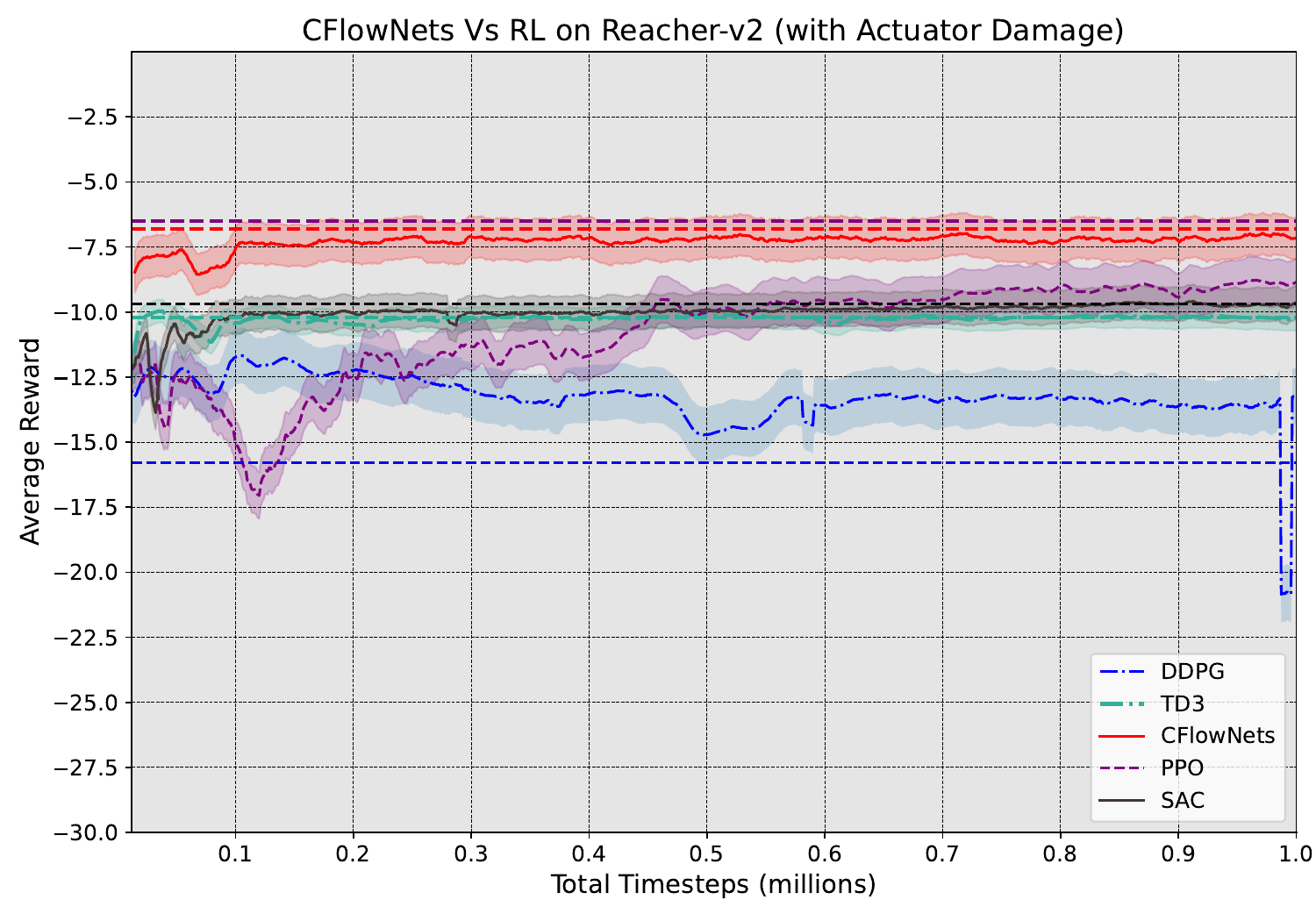}
        \caption{Actuator Damage (Fault 3)}
        \label{fig:c_act}
    \end{subfigure}
    
    \vspace{0.5cm} 
    
    \begin{subfigure}[b]{0.8\textwidth} 
        \centering
        \includegraphics[width=\textwidth]{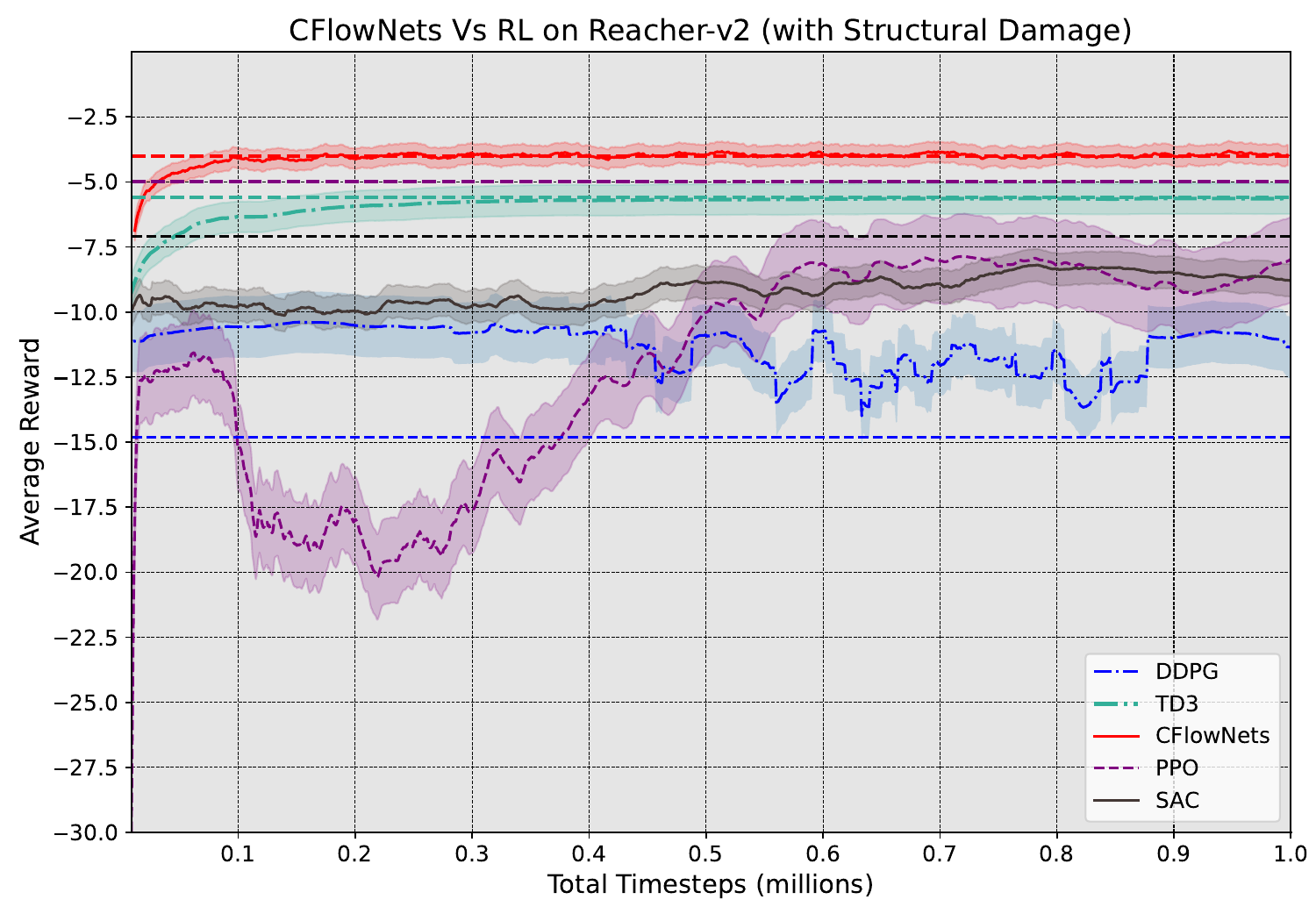}
        \caption{Structural Damage (Fault 4)}
        \label{fig:d_strut}
    \end{subfigure}
    
    \caption{The early performance in Structural and Mechanical Fault environments is depicted through learning curves for all five algorithms. The dashed line represents the asymptotic performance.}
    \label{fig:cd}
\end{figure}

\paragraph{\textbf{Structural and Mechanical Faults ((Fault 3 and Fault 4))}}
Figure~\ref{fig:cd} shows the performance of CFlowNets and the baselines 
when encountering Fault 3 and Fault 4.
We observe that CFlowNets significantly outperforms the baseline algorithms in both of the faults.
It required the least amount of experience to reach near-asymptotic performance, and compared to RL algorithms, CFlowNets showed a relatively consistent and stable learning trend.
Compared to other fault environments, CFlowNets’s performance for fault 3 was relatively unstable. 
During the initial time steps, noticeable, yet minor, fluctuations marked the learning curve 
before it plateaued post 1.2 million time steps. 
CFlowNets also showed minimal performance deviations. 

PPO, on the other hand, was susceptible to a major performance dip initially and 
was subject to a fluctuating learning trajectory throughout the predominant phase of its learning cycle 
in both fault settings. 
Despite this, PPO showed remarkable adaptability to both adversarial conditions 
by outperforming every algorithm in terms of higher asymptotic performance for the Fault 3 environment and
placing second in the Fault 4 environment. 

Although TD3 and SAC exhibited initial fluctuation 
at the early stages of their learning period for Fault 3, 
Both algorithms quickly stabilized, showing almost identical reward trends. 
However, when benchmarked against PPO and CFlowNets, 
the asymptotic performance of both TD3 and SAC was observed to be inferior. 
For the environment with structural damage, 
TD3 displayed an improved and consistent performance 
with a significantly reduced sample experience. 
Conversely, the SAC’s performance demonstrated fluctuations, 
indicating a lack of stability in the learning process. 
In the end, convergence was achieved after a lengthy learning process.

Finally, DDPG exhibited suboptimal adaptive performance 
both in terms of average reward trend and algorithmic stability. 
From Figures \ref{fig:c_act} and \ref{fig:d_strut}, 
we can see that the initial learning curve was relatively stable, 
but as the time steps advanced, a stark deviation was noted. 
For the entirety of the learning period, 
DDPG experienced multiple abrupt declines in performance. 
The algorithm’s policy proved to be inadequately robust to recover from it, 
resulting in a poor average reward gain.

\paragraph{\textbf{\textcolor{red}{Adaptation Performance (asymptotic) Across Fault Environments}}}
\textcolor{red}{
To capture the fault adaptation efficacy of each algorithm, 
we measure the asymptotic performance
(represented as dashed lines in Figures \ref{fig:ab} and \ref{fig:cd}) 
of each of the five algorithms under fault conditions. 
The asymptotic performance for each algorithm was calculated by analyzing the learning curve over the final 200,000 timesteps, 
corresponding to the last 20 evaluation rollouts.
Within this window size, 
the variance of the average rewards across evaluations was computed. 
When the variance dropped below a predefined minimal threshold, 
it was considered the stabilization point.
The asymptotic adaptation performances were calculated as 
the average return over the final 20 policy evaluation rollouts, 
with these averages further computed across 10 independent runs for greater reliability. 
These dashed lines point to each algorithm’s stability and adaptation in reaching a stable state of performance and 
the highest average reward achieved under fault conditions. 
The table \ref{table:asymptotic_performance} presents the asymptotic performance values across the four fault environments, 
highlighting each algorithm’s peak stability and adaptability in response to specific faults. 
Based on the adaptation performance results, 
CFlowNets and PPO demonstrate a greater degree of resilience and average rewards in most scenarios. 
}

\begin{table}[htbp]
    \footnotesize
    \centering
    \caption{\textcolor{red}{Asymptotic Adaptation Performance of Each Algorithm Across Fault Environments}}
    \label{table:asymptotic_performance}
    \resizebox{\textwidth}{!}{%
    \begin{tabular}{|>{\centering\arraybackslash}m{1.7cm}|>{\centering\arraybackslash}m{3cm}|>{\centering\arraybackslash}m{2.5cm}|>{\centering\arraybackslash}m{2.5cm}|>{\centering\arraybackslash}m{2.5cm}|}
        \hline
        \textbf{Algorithm} & \textbf{Reduced Range of Motion} & \textbf{Increased Damping} & \textbf{Actuator Damage} & \textbf{Structural Damage} \\
        \hline
        CFlowNets & -4.6 & -4.9 & -6.8 & -3.9 \\
        TD3       & -7.0 & -7.1 & -10.2 & -5.6 \\
        SAC       & -6.2 & -12.2 & -9.7 & -7.1 \\
        DDPG      & -13.5 & -14.1 & -15.8 & -14.8 \\
        PPO       & -4.9 & -4.7 & -6.5 & -4.9 \\
        \hline
    \end{tabular}%
    }
\end{table}

\subsection{Adaptation Speed and Sample Efficiency}
\begin{figure}[htbp]
    \centering
    \includegraphics[width=\textwidth]{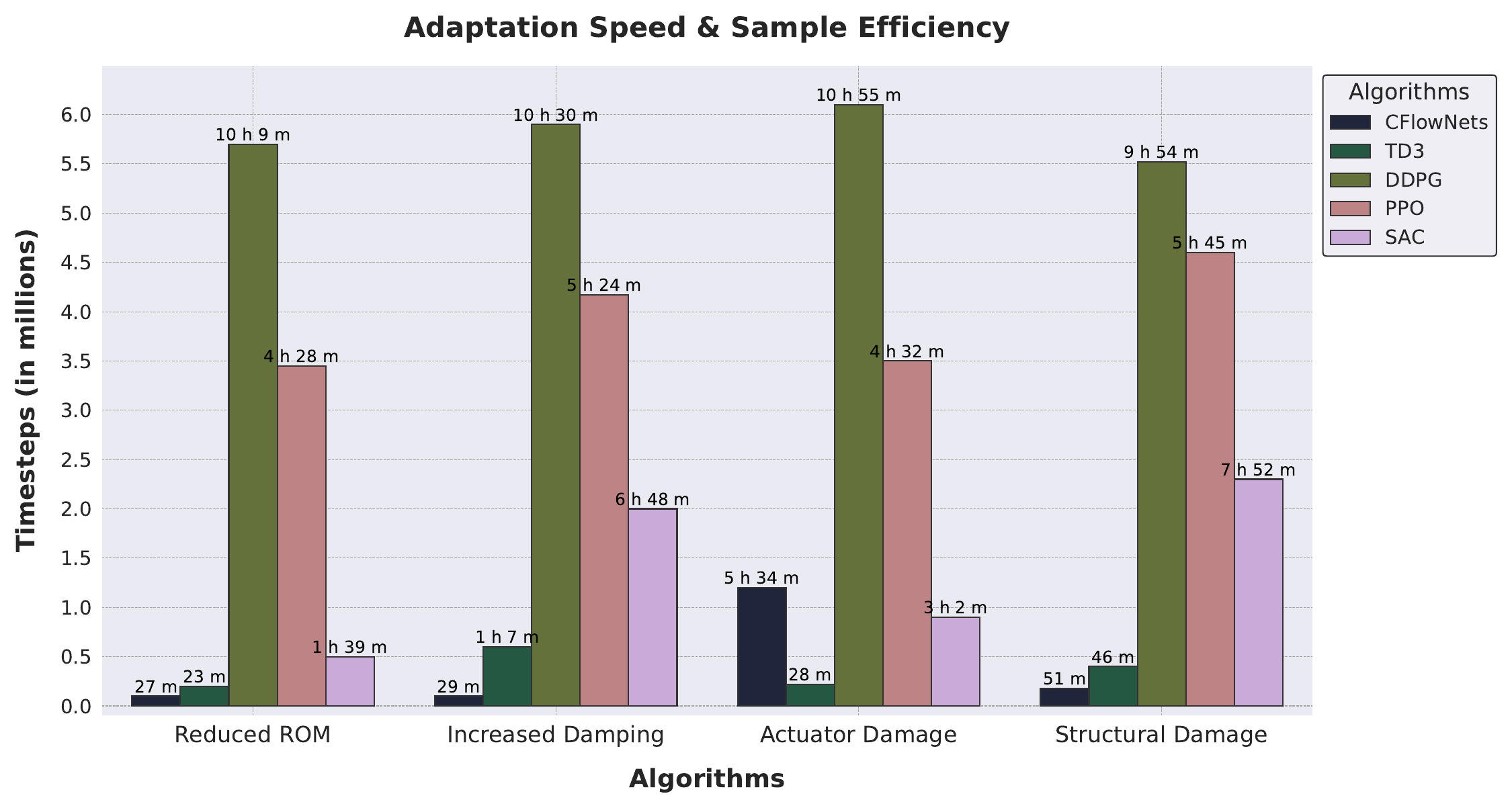}
    \caption{Adaptation Speed and Sample Efficiency in the four fault Reacher-v2 Environments (10 million time steps). Execution time to achieve asymptotic performance is indicated on top of each algorithm's bar in an Hour-Minute format.}
    \label{fig:adaptspeed}
\end{figure}

We utilize a grouped bar chart (shown in Figure \ref{fig:adaptspeed}) 
to gain more insights into each algorithm’s efficacy in the four fault environments. 
In the graph, we organize our collected data from previous experiments into four groups, 
each corresponding to one of the fault environments. 
\textcolor{red}{
The figure quantifies the time steps required for each algorithm 
to reach a point where the average reward stabilizes over a defined number of episodes or where the performance curve shows minimal fluctuations (asymptotic performance). 
}
To understand adaptation speed better, we incorporate on top of the corresponding bar, 
the real-time duration required for each algorithm
to converge in an hour-minute format.

From Figure \ref{fig:adaptspeed}, 
we can see that CFlowNets demonstrated remarkable performance 
both in terms of fast adaptation and sample efficiency for most of the fault environments. 
The model took between 27 to 51 minutes 
and approximately only 100k -- 200k time steps 
to gain asymptotic performance for Faults 1, 2 and 4. 
However, a noticeable anomaly is exhibited in Fault 3 (actuator damage) 
where CFlowNets policy required more than a million time steps 
(approx. 1.2 million) 
with an execution time of 5 hours and 34 minutes to converge. 

TD3, on the other hand, 
had one of the best performances 
in terms of requiring fewer real experiences (300k -- 700k time steps) 
and lower execution time to adapt. 
Although its performance did not surpass CFlowNets in the majority of the environments, 
it outperformed all the rest of the three RL models in terms of adaptation speed.

The performance of SAC was not consistent across the four environments.
It excels in the Reduced Range of Motion (Fault 1) and Actuator Damage (Fault 3) environments 
but encounters challenges in Increased Damping (Fault 2) and Structural Damage (Fault 4), 
taking a relatively greater number of samples before it converges.
Due to the different dynamics and nature of the fault environments, SAC policies might be adequate to handle certain complexities while being sensitive to others.

PPO experiences significant challenges in terms of sample efficiency
due to its on-policy nature.
From the grouped bar chart in Figure~\ref{fig:adaptspeed}, 
it is evident that PPO necessitates much more real experiences 
ranging from 3.45 to 4.6 million time steps, hence the longer execution time requirement, 
before it can adapt to the malfunctions. 
This low sample efficiency may undermine its applicability in tasks where sample efficiency and faster adaptation are paramount considerations, 
such as search and rescue applications~\cite{niroui2019deep}.

Finally, it can be easily seen from Figure~\ref{fig:adaptspeed} that 
DDPG requires significantly more time than the other algorithms 
(approximately 5.5 to 6.1 million time steps) and 
execution time running between 9 to 11 hours 
to plateau in its performance curve. 
Similar to PPO, DDPG has limited applicability in scenarios requiring rapid adaptation.

\textcolor{red}{
To provide a comprehensive analysis of each algorithm's adaptation speed and sample efficiency across all fault environments, 
Table \ref{table:combined_adaptation_speed} presents the timesteps and 
real-time execution hours required for each algorithm to reach asymptotic performance. 
The Timesteps column quantifies the number of experiences each algorithm needed to stabilize, 
measured in millions. 
The Time column represents execution time in hours and minutes to converge. 
This table highlights the relative speed of adaptation and computational demand for each algorithm and 
offer valuable insights into their applicability real-time fault adaptation tasks. 
Based on this table, we observe that algorithms like CFlowNets and TD3 demonstrate fast convergence in most environments, 
whereas DDPG and PPO require significantly more samples and time to reach stability.
}

\begin{table}[htbp]
    \centering
    \caption{\textcolor{red}{Comparative Convergence Analysis: Timesteps and Execution Time per Fault Environment}}
    \label{table:combined_adaptation_speed}
    \resizebox{\textwidth}{!}{%
    \begin{tabular}{|l|cc|cc|cc|cc|}
        \hline
        \textbf{Algorithm} & \multicolumn{2}{c|}{\textbf{Reduced Range of Motion}} & \multicolumn{2}{c|}{\textbf{Increased Damping}} & \multicolumn{2}{c|}{\textbf{Actuator Damage}} & \multicolumn{2}{c|}{\textbf{Structural Damage}} \\
        & \textbf{Timesteps} & \textbf{Time} & \textbf{Timesteps} & \textbf{Time} & \textbf{Timesteps} & \textbf{Time} & \textbf{Timesteps} & \textbf{Time} \\
        & \textbf{(million)} & \textbf{(hrs)} & \textbf{(million)} & \textbf{(hrs)} & \textbf{(million)} & \textbf{(hrs)} & \textbf{(million)} & \textbf{(hrs)} \\
        \hline
        CFlowNets & 0.10 & 0h 27m & 0.10 & 0h 29m & 1.20 & 5h 34m & 0.18 & 0h 51m \\
        TD3       & 0.20 & 0h 23m & 0.60 & 1h 7m  & 0.22 & 0h 28m & 0.40 & 0h 46m \\
        DDPG      & 5.70 & 10h 9m & 5.90 & 10h 30m & 6.10 & 10h 55m & 5.52 & 9h 54m \\
        PPO       & 3.45 & 4h 28m & 4.17 & 5h 24m & 3.50 & 4h 32m & 4.60 & 5h 45m \\
        SAC       & 0.50 & 1h 39m & 2.00 & 6h 48m & 0.90 & 3h 2m & 2.30 & 7h 52m \\
        \hline
    \end{tabular}%
    }
\end{table}

\subsection{Execution Time}
\begin{figure}[htp]
    \centering
    \includegraphics[width=\textwidth]{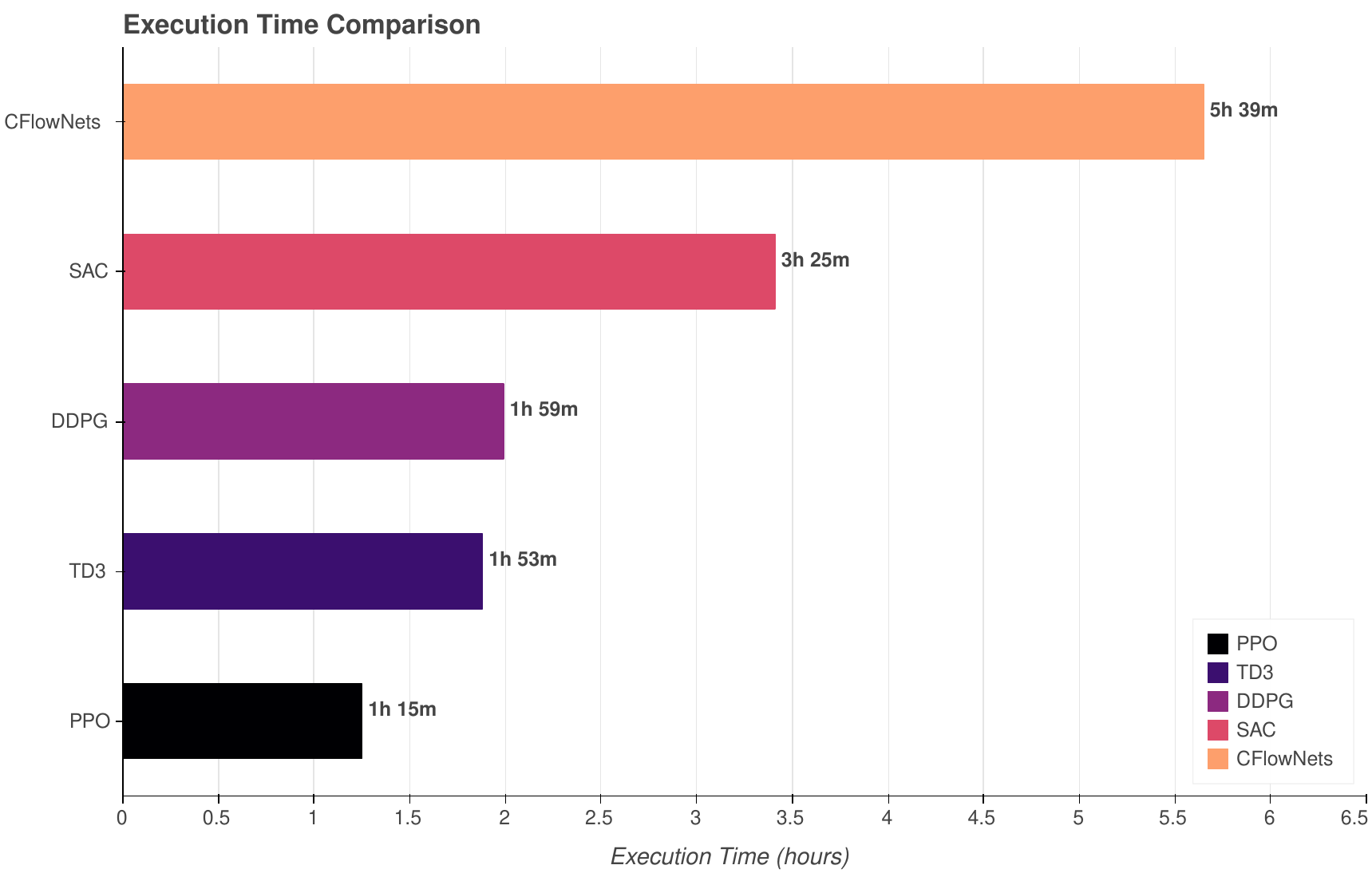}
    \caption{Execution Time in Normal Reacher-v2 Environment (1 million time step).}
    \label{fig:execution(ne)}
\end{figure}

In this segment,
we analyze the execution time of the algorithms in Stage 1.
Figure~\ref{fig:execution(ne)} shows the required time by CFlowNets and our baseline algorithms
to run for 1 million time steps in a normal robot environment (Stage 1).
This is a part of our preliminary experiment to get initial insights into the performance of CFlowNets. 
It helps to establish an initial benchmark and is not indicative of the optimal operation for each algorithm.


From Figure \ref{fig:execution(ne)}, 
we see that CFlowNets has the longest execution time (5 hours and 39 minutes).
In comparison, other RL methods require significantly less amount of time, 
roughly between 1 to 3.5 hours. 
This is due to the fact that CFlowNets generates a distribution over all possible paths, 
and samples from the most rewarding paths with a higher probability. 
Although this feature enables CFlowNets to showcase superior performance in terms of accumulating rewards, 
is comes at a computational cost. 
The continuous normalizing flows, which empower CFlowNets in modeling complex dynamics, 
is inherently computationally intensive. 
It is also worth noting that the reported execution time 
only accounts for the training of flow networks 
which already have a pre-trained Retrieval network.

\subsection{CFlowNets' Compute Efficiency Analysis: GPU Memory Usage}
\begin{figure}[H]
    \centering
    \includegraphics[width=\textwidth]{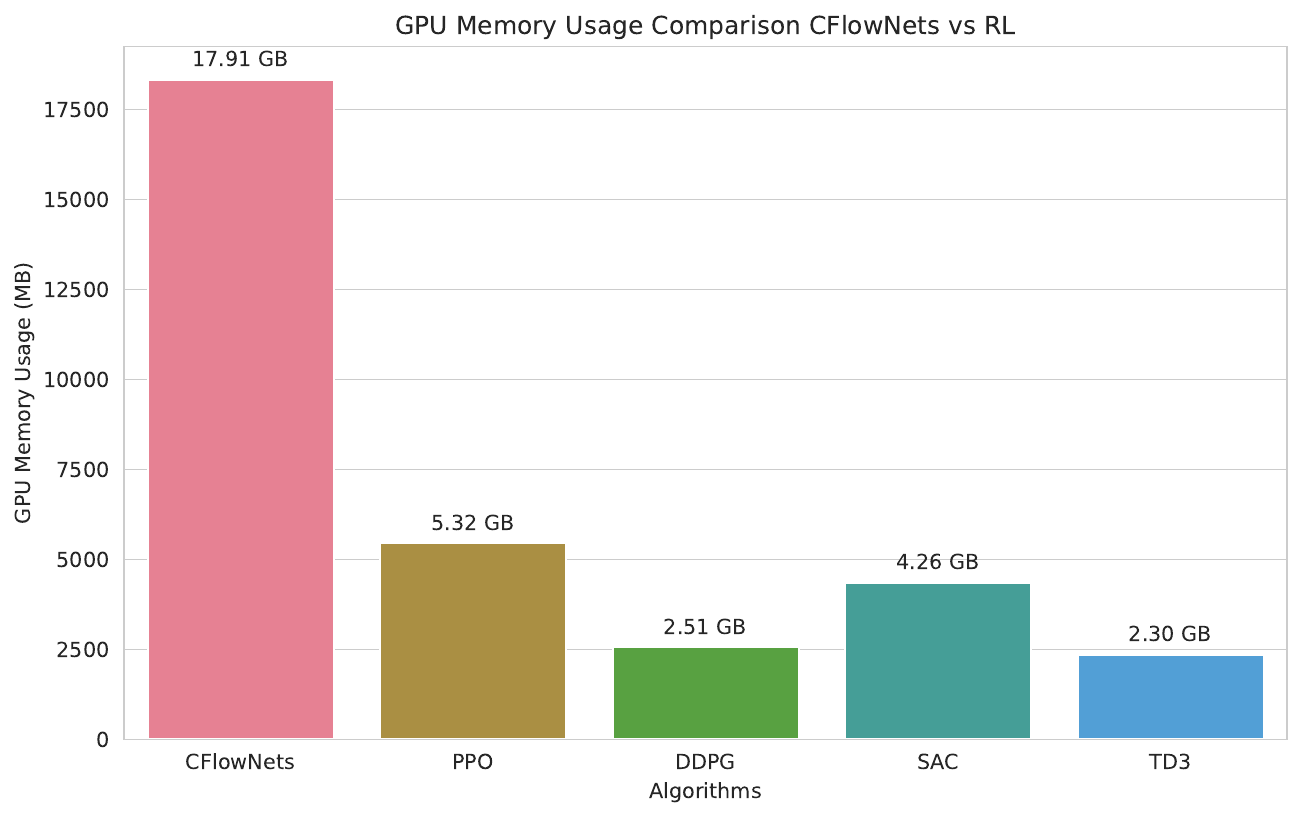} 
    \caption{Bar chart illustrating the average GPU memory usage of various algorithms: CFlowNets, PPO, DDPG, SAC, and TD3 for Reacher-v2 (Normal Environment)}
    \label{fig:compute}
\end{figure}

This segment explore the compute resource usage of CFlownets compared to the four RL algorithms. 
These algorithms' operational efficiency during the training phase, 
is crucial to their applicability in real-world machine fault adaptation scenarios. 
The data collected for this comparison 
is the GPU memory usage of each algorithm in a normal Reacher-v2 algorithm. 
The GPU memory usage data was collected at regular intervals. 
we report the average of the collected data for each evaluation time step in Figure~\ref{fig:compute}.
According to this figure, 
CFlowNets consumed on average a substantial 17.91 GB of GPU memory during its execution.
In contrast, the RL algorithms necessitated less than one-third of the GPU memory to perform the same task.
Model-free RL algorithms such as PPO, SAC, DDPG, and TD3 
learn a policy that directly maps states to actions 
while sticking to the most rewarding paths.
On the other hand, CFlowNets, due to its sampling strategy,
explores the solution space more exhaustively.
Thus, it is bound to store and process a significant amount of data 
corresponding to all the possible paths and their associated states and actions.
As a result, CFlowNets requires much more computation resources 
than standard reinforcement learning methods.

\subsection{CFlowNets’s Model and Storage Transfer Analysis}
In our final segment of the comparative evaluation, 
we further analyze how transfer learning affect the performance of CFlowNets
by transferring both the model parameters and storage contents
obtained in Stage 1.
As we observed the superior performance of CFlowNets in the first segment,
we do not include other RL algorithms for this experiment.


In this experiment, 
we train CFlowNets in the original Reacher-v2 normal environment (Stage 1). 
We save the model parameters learned in the normal environment 
along with the replay buffer 
and then transferred to the four fault environments. 
We evaluate the performance of the CFlowNets, 
when both model parameters and replay buffer are retained,
and compare it with when only the model parameters are retained.


\begin{figure}[p]
    \centering
    
    \begin{subfigure}[b]{0.8\textwidth}
        \centering
        \includegraphics[width=\textwidth]{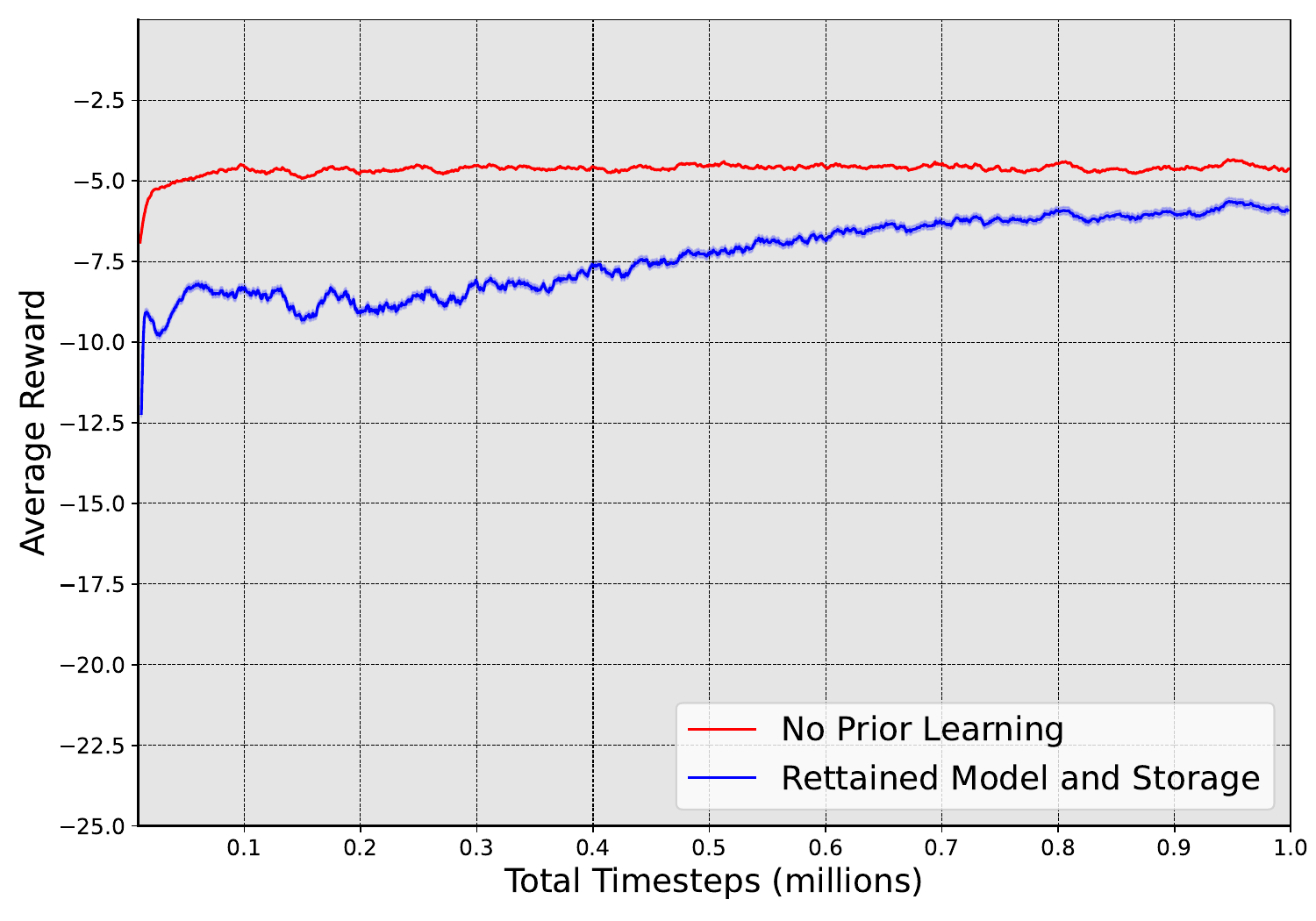}
        \caption{Reduced ROM}
        \label{fig:a_ROM_}
    \end{subfigure}
    
    \vspace{0.3cm}
    
    \begin{subfigure}[b]{0.8\textwidth}
        \centering
        \includegraphics[width=\textwidth]{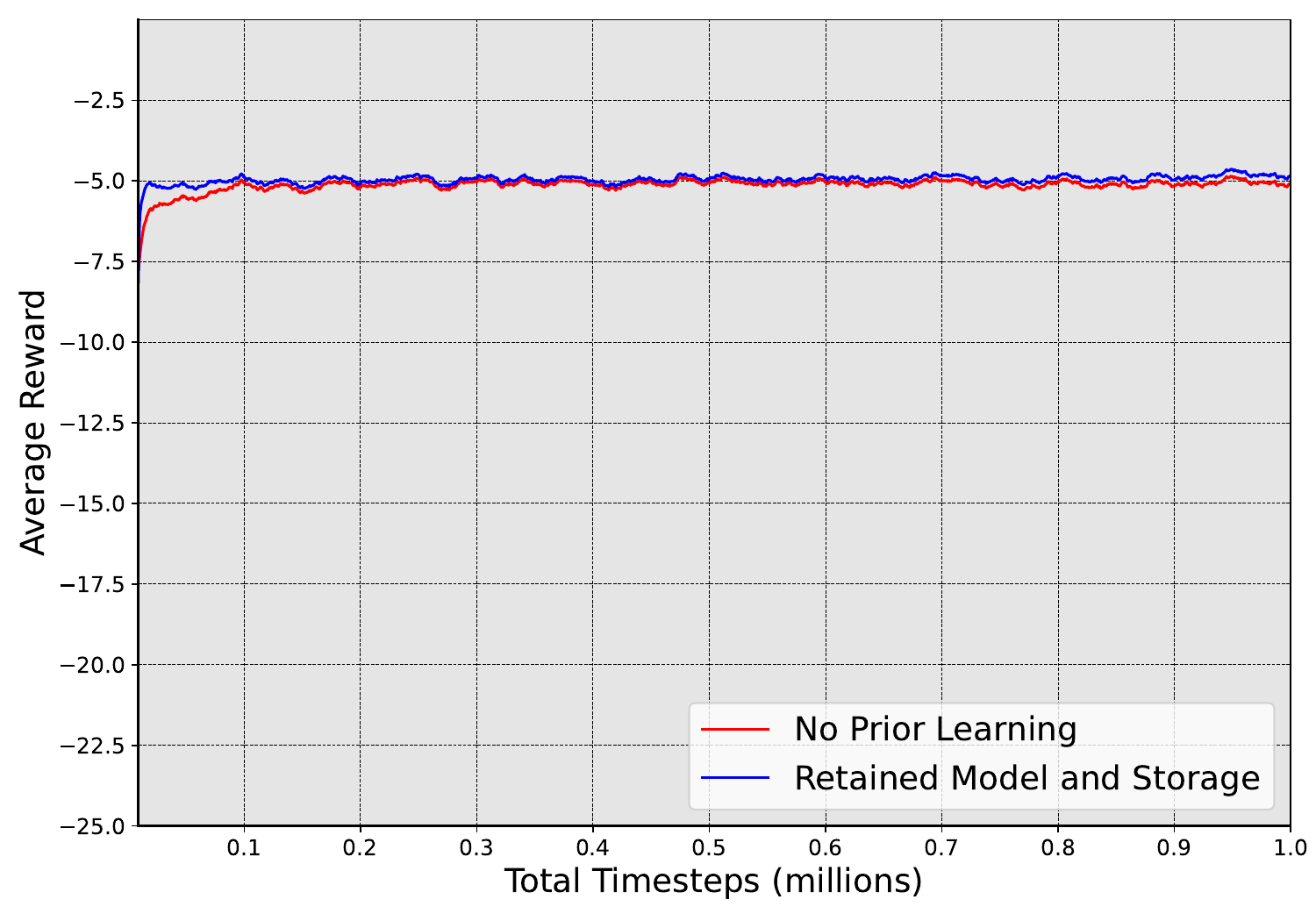}
        \caption{Increased Damping}
        \label{fig:b_Damping_}
    \end{subfigure}
    
\end{figure}

\begin{figure}[p]
    \ContinuedFloat 
    \centering

    \begin{subfigure}[b]{0.8\textwidth}
        \centering
        \includegraphics[width=\textwidth]{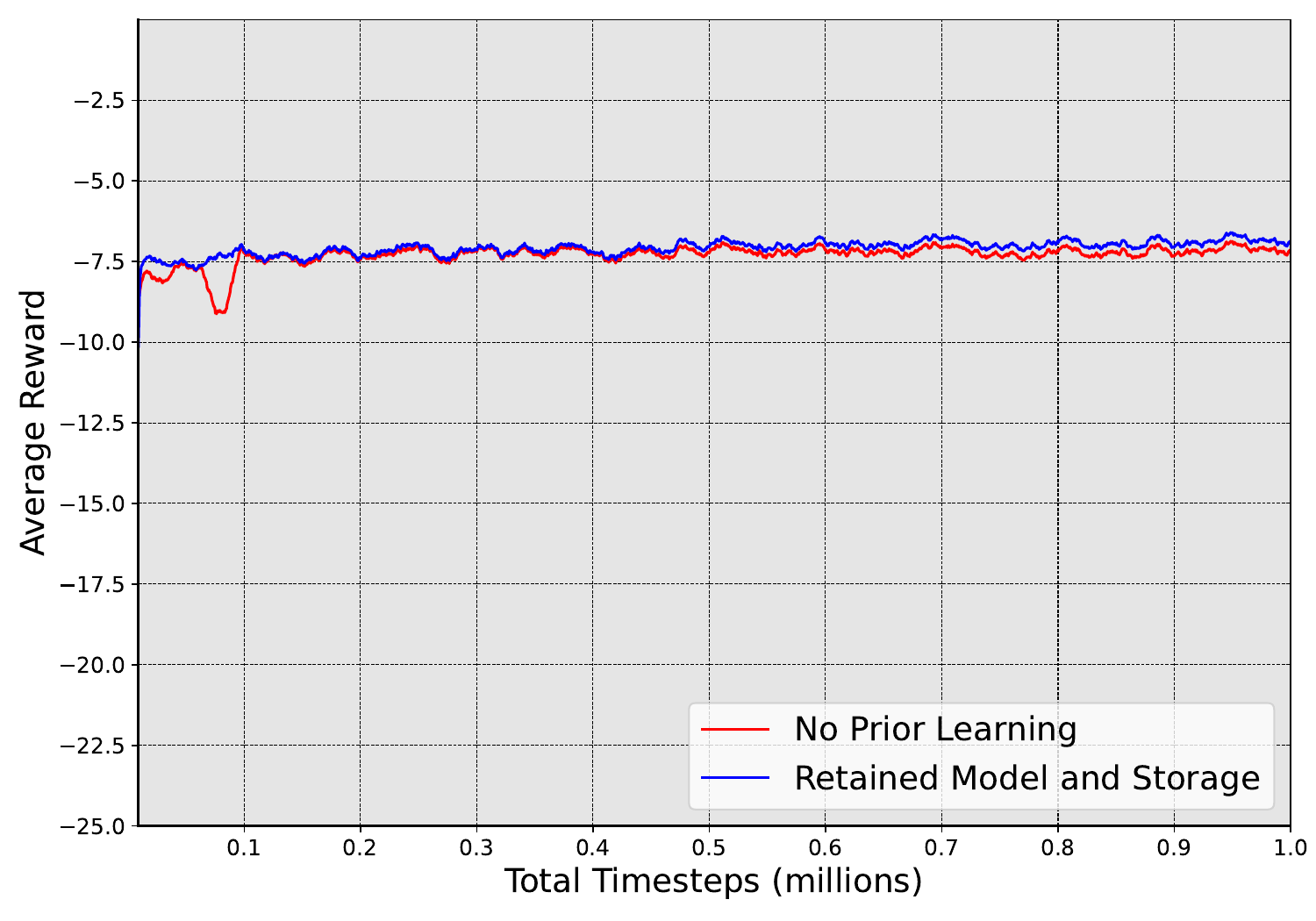}
        \caption{Actuator Damage}
        \label{fig:c_description}
    \end{subfigure}
    
    \vspace{0.3cm}
    
    \begin{subfigure}[b]{0.8\textwidth}
        \centering
        \includegraphics[width=\textwidth]{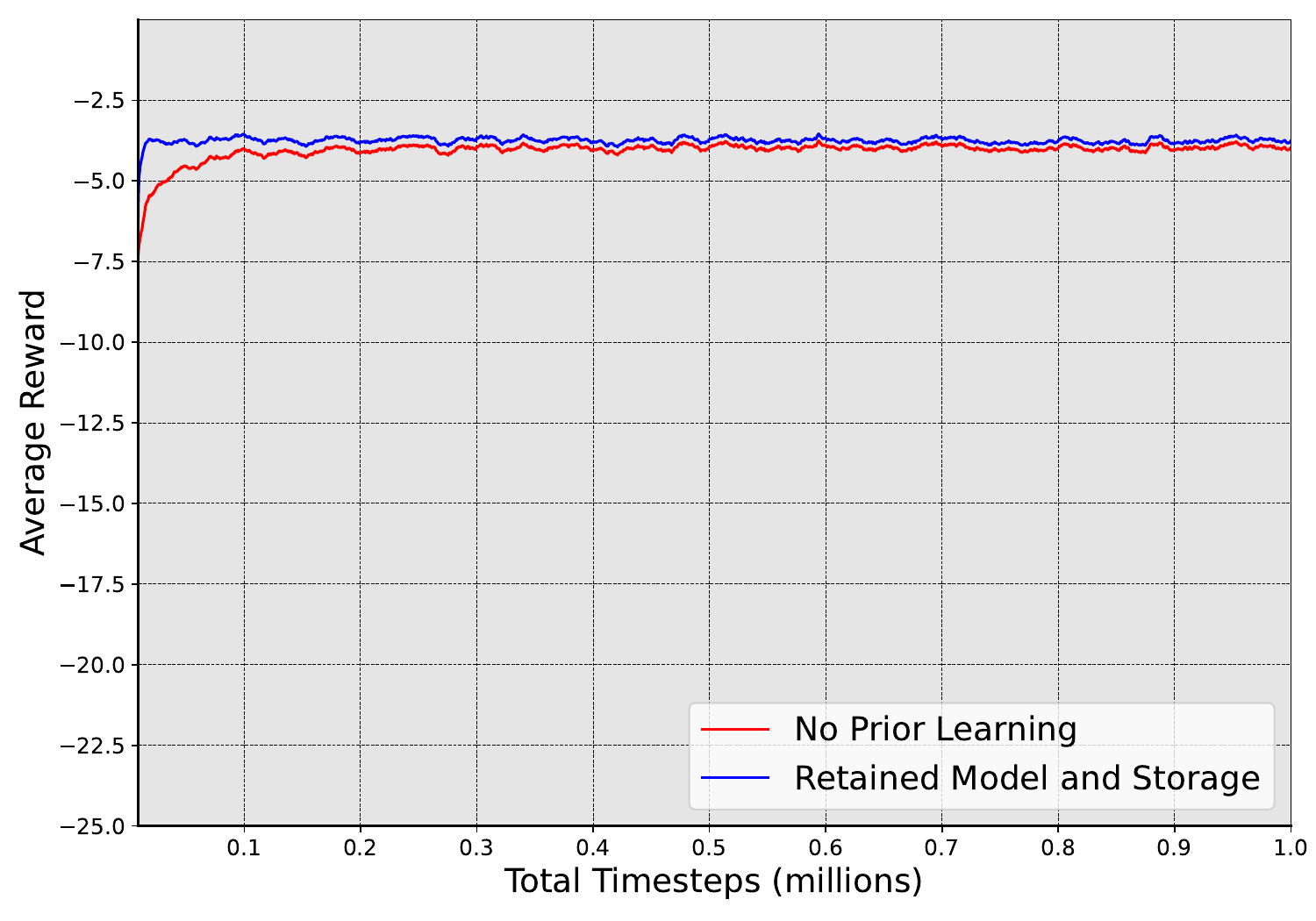}
        \caption{Structural Damage}
        \label{fig:d_struct}
    \end{subfigure}
    
    \caption{Early Performance of CFlowNets in Reacher-v2 fault environments. (a) Reduced ROM, (b) Increased Damping, (c) Actuator Damage, and (d) Structural Damage. Comparison of performance is done between the CFlowNets model with no prior learning and the CFlowNets model with retaining pre-trained model and replay buffer (data collected from the normal Reacher-v2 environment).}
    \label{fig:task_transfer}
\end{figure}

Figure~\ref{fig:a_ROM_} illustrates that 
transferring the model parameters and the replay buffer contents of the normal environment 
do not significantly enhance the CFlowNets' performance; 
rather, for the environment with Fault 1, 
the performance was worse than when retaining only the model parameters. 
One of the potential reasons for this observation could be 
the mismatch of experience relevance, 
as the experiences collected in the normal environment may have little relevance 
to the altered environment with a reduced range of motion 
where the dynamics of the environment are drastically different from the original environment. 
Consequently, the learned policies and the past experience, 
including the action and state transition, 
may no longer be valid and counterproductive for the new task, 
leading to a sub-optimal performance for Fault 1.

Conversely, when examining Figures~\ref{fig:b_Damping_}, 
\ref{fig:c_description}, and \ref{fig:d_struct}, 
it can be seen that across the rest of the three fault environments,
retaining both model parameters and storage contents in CFlowNets
leads to better and more stable performance
compared to only retaining model parameters,
especially during the initial 100,000 time steps.
To explain this, notice that in our experimental setup,  
the process of collecting the experiences in the normal environment
and deploying them in a fault environment
is akin to offline learning.
In fault environments, 
such models are exposed to online and out-of-distribution experiences. 
When we retain the pre-trained models and the storage contents in the post-fault tasks, 
the impact of encountering online and out-of-distribution experiences 
is substantially reduced in the beginning. 
That is why when we transfer knowledge from Stage 1 (source task) to the modified target tasks 
it improves the initial performance of the agent 
when it starts learning a new task using knowledge learned from a previous related task.
This observation is often referred as jumpstart~\cite{jmlr09-taylor}. 
When we transfer the pre-trained models along with experiences 
from the normal environment to the fault environment, 
it mitigates the cold start 
because the agent already has a good estimate for the new environment 
by utilizing past experiences. 
Instead of learning from scratch, 
the replay buffer enables the agent to exploit good policies 
from the start without random exploration.

In later time steps, 
the model converges to a similar learning curve 
as the model with only retained model parameters 
because of the buffer saturation. 
As the finite replay buffer is constantly being populated 
with new experiences from the current environment, 
it overwrites the old ones. 
Consequently, as the agent gathers more samples from faulty environments, 
the old experiences may become less useful.

\section{Discussion}
Throughout the series of experiments, 
we demonstrated the applicability of CFlowNets 
in continuous control tasks.
We specifically showed that its algorithmic structure 
is well-suited for effective adaptation in simple robotic simulations.
We found that CFlowNets excels at tasks requiring fast adaptation 
with the least amount of experience.
Nevertheless, its significant resource consumption 
poses certain challenges in applications with limited resources.
Although deploying machine learning models is a common challenge in production,
and it is an open research question~\cite{marculescu2018hardware,paleyes2022challenges},
future work should investigate different methodologies to optimize CFlowNets
for real-world applications, especially those with limited resources.

Among the RL algorithms, PPO also demonstrated commendable adaptation performance. PPO’s average reward trajectory converged higher than CFlowNets in some fault environments. Nevertheless, it required a large number of experiences before stabilizing. Thus, the algorithm took significantly longer to adapt in all the fault environments compared to CFlowNets.
Given the fact that PPO requires significantly less execution time and GPU memory,
it can be an ideal choice for robotic applications with limited resources,
but sacrificing adaptation speed.

Based on our findings, CFlowNets and PPO emerged as commendable top-tier performing algorithms 
for our fault adaption task. 
We further analyze CFlowNets and PPO in terms of their performance retention,
which is crucial to understanding how algorithms adjust their policies 
in response to adversarial conditions to maintain their asymptotic performance.
Therefore, we determine what percentage of normal performance CFlowNets and PPO can retain. 
To compute this, we utilize the point of asymptotic performance across 10 million time steps 
for CFlowNets and PPO in the normal environment and 
compare them with the asymptotic performance in the four fault environments. 
Overall, CFlowNets can retain 68.43\% to 94.74\% of its normal performance 
in Fault 1, Fault 2, and Fault 4 environments. 
However, in Fault 3, a substantial reduction of 78.95\% is observed. 
This is because the actuator damage introduces a more complex non-linearity in the environment. 
Conversely, for Fault 3, PPO also experienced a 59.54\% asymptotic performance reduction. 
While this decrease is less severe than CFlowNets, 
it is a considerable reduction. 
For the rest of the fault environments, 
PPO was able to retain 78.05\% to 85.37\% of its original performance. 
This evaluation provides insight into the varied impact of each fault condition 
on the two algorithm’s asymptotic performance. 
CFlowNets demonstrated a higher degree of resilience under specific fault environments 
such as Fault 4, 
whereas PPO maintains a more stable performance retention in others.

A potential reason for CFlowNets’ superior performance is
its trajectory sampling strategy, i.e., 
it generates a distribution over all possible paths and 
samples from the most rewarding paths with a higher probability. 
CFlowNets’ ability to generate a distribution over trajectory means that 
it can explore multiple potential paths concurrently. 
As a result, the possibility of the algorithm getting stuck in a local optimum reduces 
and makes it more suitable to find more globally optimal solutions, 
which makes them sample efficient. 
By prioritizing the most rewarding paths, 
CFlowNets' sampling mechanism continuously fine-tunes its parameters based on the highest returns. 
As a result of this prioritization, convergence occurs faster and results in more optimal performance.
Additionally, because of its off-policy nature, 
it can utilize stored experiences to learn effectively from a limited number of samples 
by reusing experiences multiple times. 
In continuous state and action space, 
CFlowNets' architecture may facilitate more accurate state generalization. 
In this way, it is capable of recognizing and acting optimally in unseen but similar states, 
which is invaluable in dynamic environments.

On the other hand, for every fault environment, 
PPO faces a performance dip initially in its learning curve, 
and it requires a lot of time steps to reach asymptotic performance. 
First and foremost, PPO is an on-policy algorithm 
thus its learning phase depends on the most recent experiences. 
However, initially, these experiences are based on random or suboptimal policies. 
This on-policy feature may be beneficial in the long run, 
but it requires more time steps to collect valid experiences 
that contribute to optimal policy improvements~\cite{2021arXiv211100072Q}. 
Additionally, PPO utilizes a clipping mechanism in its objective function, 
which hinders large policy updates. 
This strategy prevents PPOs from overshooting policy updates 
and ensures training stability.
However, this might cause a slower rate of improvement in policy updates during the early phases of training.
Therefore, the algorithm takes a long time to reach asymptotic performance.

\textcolor{red}{
In the context of sample efficiency, 
TD3 demonstrated similar performance to CFlowNets 
in all the fault environments illustrated in Figure~\ref{fig:ab} and Figure~\ref{fig:cd}. 
However, despite its good sample efficiency, 
the overall average reward trajectory was lower compared to CFlowNets and PPO. 
The reason might be due to the fact that 
TD3 utilizes a twin-value network to estimate the Q-values and delayed policy updates 
to reduce overestimation bias and ensure a more stable training phase, respectively. 
Additionally, TD3 uses target policy smoothing, which adds noise to the target policy, 
making the policy more robust and the estimation of Q-values more conservative. 
The delayed policy updates and target policy smoothing may result in efficient learning, 
but it also creates an obstacle to exploration. 
As a result, TD3 may produce lower rewards than other methods, such as CFlowNets and PPO. 
In addition to this, TD3 is an off-policy algorithm that learns from a replay buffer. 
Although more sample efficient, 
this approach might cause the model to stuck in outdated and suboptimal policies, 
causing a lower average reward trajectory.
}

\textcolor{red}{
Moreover, SAC (Soft Actor-Critic) exhibited a stable average reward trajectory 
for most of the experiments in the four custom gym environments. 
However, if we compare its performance with CFlowNets, PPO, and TD3, 
we observe that its average reward trajectory was lower. 
Despite this poor adaptive performance, one notable feature is its good exploration ability. 
In the context of sample efficiency, 
SAC was able to quickly achieve convergence for most of the fault environments.
SAC is entropy-regularized, meaning that it uses an entropy term to encourage more exploration. 
As a result, SAC maintains a more stochastic policy and explores more in the environment, 
explaining its good sample efficiency. 
Furthermore, due to its off-policy nature, 
SAC is more sample-efficient and requires fewer timesteps to achieve near-asymptotic performance. 
This may, however, lead to the algorithm being less responsive to recent changes 
than on-policy algorithms such as PPO. 
Therefore, this can be an underlying reason behind SAC’s lower accumulation of reward.
}

\textcolor{red}{
Among the RL algorithms, 
DDPG showed the lowest adaptive performance over 10 million timesteps, 
demonstrating considerable limitations in sample efficiency, 
reward accumulation, 
and adaptation speed. 
There was fluctuation in average reward across all four fault environments, 
indicating a lack of adaptability. 
Apart from demonstrating instability, 
the results depicted a major abrupt dip in its performance. 
There could be various potential reasons why DDPG is not suitable for fault adaptation tasks. 
To further analyze these observations, 
we examine DDPG's performance with findings reported in similar studies 
that implemented DDPG in Reacher or comparable robotic environments.
Lillicrap et al. \cite{lillicrap2015continuous} introduced DDPG as an actor-critic method 
specifically designed for continuous control tasks. 
They implemented the algorithm in a number of robotic environments including Reacher. 
Their results indicated that DDPG could achieve stable performance in standard setups, 
but its performance is highly sensitive to hyperparameters, 
especially in high-dimensional action spaces. 
These insights correctly align with our observations that 
DDPG's effectiveness was constrained in fault-adaptation tasks 
as we followed published hyperparameters with selective exploration tuning. 
This sensitivity to hyperparameters has also been reported in another study conducted 
by Henderson et al. \cite{henderson2017deep}.
where they investigated the reproducibility challenges of DDPG in simulated environments. 
Their results confirmed that DDPG's performance can vary to a higher degree without thorough hyperparameter tuning. 
Even minimal variation in settings led to inconsistent performance, demonstrating the sensitivity of DDPG to initial configurations and hyperparameter optimization. 
In our experiment, we followed published hyperparameters with selective exploration tuning; hence, DDPG may have behaved inconsistently in our experiments across different fault scenarios. 
Therefore, it is evident that for tasks requiring fault adaptation, 
thorough hyperparameter search is necessary. 
In addition, Fujimoto et al. \cite{fujimoto2018addressing} expanded on DDPG's limitations and 
identified that the algorithm has a tendency to cause function approximation errors and 
overestimation bias of the Q-values of the critic network. 
This can cause the agent to get stuck into a local optimum because of suboptimal policy updates.
They demonstrated that these biases, often exacerbated in dynamic and noisy environments, 
led to suboptimal policy updates. 
These findings align with the observations in our experiments, wherein 
non-linearities and noise were introduced in the faulty environments. 
These comparative studies suggest that while DDPG can perform effectively in stable robotic environments, 
fault-adaptive tasks impose additional demands on sample efficiency and stability.
}

\textcolor{red}{
In our experiments, CFlowNets consistently demonstrated faster adaptation and higher sample efficiency across multiple fault environments compared to traditional RL algorithms. 
However, these performance improvements come with notable trade-offs in terms of execution time and memory usage. 
CFlowNets require substantial GPU memory which can be resource-intensive, 
limiting their applicability in specific real-world scenarios where resources are constrained. 
It is crucial to consider the scalability of CFlowNets in more complex environments or when dealing with larger datasets. 
For example, in real-world robotic applications, as the state-action space expands in high-dimensional robotic tasks, 
the resource demands of CFlowNets are expected to increase substantially. 
Some hardware designs may require minimal resources which may not be able to support CFlowNets’s memory-intensive computations. 
In real-world tasks with more complex environments, 
the algorithm may not scale well and may face deployment issues. 
Given the fact that RL algorithms, such as PPO, require significantly less execution time and GPU memory, 
it can be an ideal alternative for robotic applications with limited resources, 
but sacrificing adaptation speed. 
From our experiments, we find that CFlowNets may be better suited for server-based systems where resources are not constrained,
and faster adaptation is required using a minimal number of samples. 
In contrast, state-of-the-art RL algorithms like PPO may be more practical in embedded systems and resource-constrained scenarios (IoT devices for automation), because of their low computational requirements.  
}

\textcolor{red}{One of the limitations of our study is that we conducted our experiments on a 2D simulated robotic task. 
Although we created four custom gym environments, 
at its core, they are different variations of the same environment. 
This setup serves as a foundational demonstration of CFlowNets’ in fault environments.
An immediate future work can further study CFlowNets for more complex, 
3D robotics simulations (e.g. FetchReach) and larger problems, 
such as human-robot collaboration in manufacturing~\cite{matheson2019human}, and
high-dimensional robotic environments~\cite{chen2023adapt,capolei2019biomimetic}. 
Such studies could examine whether the observed performance and sample efficiency advantages 
extend to real-world conditions, helping to validate CFlowNets as a robust approach for fault adaptation 
across a broader range of robotic applications.}

It is crucial to mention that
CFlowNets had an added advantage due to its pre-trained retrieval network, 
which resulted in the agent having a better estimate of the environment’s state and action space. 
This component 
might be a reason why CFlowNets was able to quickly gain convergence compared to other RL algorithms.
However, it is also imperative to acknowledge the fact that 
CFlowNets operated in a sparse reward-structured environment 
whereas the RL algorithms received intermediate rewards. 
As a result, all the RL algorithms had an advantage 
because these intermediate rewards at each time step guided the learning process 
resulting in a more efficient policy update. 
On the other hand, CFlowNets experiments were done in a sparse reward setting 
where the agent received a reward only at the terminal state at the end of an episode. 
This infrequent feedback from the environment is a significant disadvantage 
that can hinder the overall adaptation performance of CFlowNets. 
Therefore, while CFlowNets' pre-trained retrieval network offered an initial advantage, 
the sparse reward setting presented a unique challenge, 
balancing the overall comparison with traditional RL algorithms. 
As a future work, we intend to further analyze how retrieval and flow networks contribute
to the adaptation speed and quality of CFlowNets in robotic tasks. Additionally, the next step in this research will be to train both the retrieval and flow networks simultaneously to make the comparison fairer.

Another key limitation of our research was that an intensive hyperparameter search was not conducted for the implementation of the RL algorithms. In most cases, published hyperparameters for the same Reacher-v2 task were used with minor tuning. We did a selective hyperparameter tuning and focused more on hyperparameters that facilitate exploration. For a fairer comparison, we should do an intensive hyperparameter search to optimize the RL algorithms in the normal environment.

In our study, we have simulated fault environments by changing attribute variables in the XML file of our robotic environment. Each of these attributes defines certain aspects of the robot. However, all of the four-fault environments had constant fault types where the values of the modified attributes remained the same over time. In real-world scenarios, a robot typically faces gradual malfunction over time where the severity of faults generally increases; for instance, the joint becomes more and more restrictive because of gradual wear and tear. Instead of sudden and abrupt malfunction to the environment, for future experiments, the fault could be simulated in such a way that the values of the attributes gradually change as the timestep progresses. As the dynamics of the environment change bit by bit, it would be interesting to investigate if CFlowNets can demonstrate better adaptability in this sort of progressive learning setting. \textcolor{red}{Additionally, in this research, we evaluated the adaptation performance of each algorithm across individual fault scenarios. However, in real-world environments, systems often encounter multiple faults simultaneously. For instance, a robotic arm can experience increased friction and a reduced torque output at the same time. In future research, we plan to explore the adaptive performance of CFlowNets in environments with coupled faults—such as custom fault environments that combine actuator damage with increased damping or structural damage.}  In our opinion, adding multiple faults to a single testing environment will certainly increase the complexity of the environment itself. However, since CFlowNets’ has superior exploration capability because of its architecture being able to generate and sample from a distribution over multiple potential trajectories, we can hypothesize that it will still demonstrate commendable performance. Evaluating the adaptation performance of CFlowNets under such multi-fault conditions will be a good addition to a more in-depth comprehensive assessment of CFlowNets' adaptive performance in dynamically complex environments.

\section{Conlcusion}
In this paper, 
we have explored the application of Continuous Flow Networks (CFlowNets) to robotic tasks.
We have designed an experimental setup where we simulate four common robotic faults.
We studied how CFlowNets perform in these fault environments and
compared it against state-of-the-art Reinforcement Learning (RL) algorithms.
Our hypothesis was that CFlowNets' sampling strategy for generating trajectories
enables efficient exploration of high-dimensional search space in robotic tasks,
especially when quick adaptation to faults is necessary.
We have conducted multiple experiments,
i.e., adaptation performance, speed, and efficiency, and execution time and resources consumption,
to obtain key insights into CFlowNets' performance in a robotic environment.

Overall, we conclude that 
CFlowNets outperforms state-of-the-art RL algorithms 
in a widely used Reacher-v2 robotic environment.
We found that CFlowNets is a top-tier algorithm 
when it comes to incorporating hardware fault adaption to the robotic arm.
Specifically, we observe that
CFlowNets achieved a high asymptotic performance, 
surpassing the state-of-the-art RL algorithms.
Across our four custom gym environments with faults, 
CFlowNets have exhibited excellent sample efficiency, 
requiring the least number of timesteps. 

In addition, we found that CFlowNets, especially with retaining model and replay buffer, 
can be considered a promising option for rapid adaptation in robots.
The use of CFlowNets for sampling distributions over complex spaces is suitable 
for tasks that require a diverse set of solutions through comprehensive exploration. 
However, for tasks that are strictly about maximizing cumulative rewards, 
traditional RL algorithms might be more suitable. 

This study is the first step in successfully applying CFlowNets
to exploration-biased robotic tasks 
with machine fault adaptation. 
This study promises further research in CFlowNets 
that will potentially lead to more reliable and efficient real-world robots in various fields, 
making them a common choice for tasks where exploration and adaptability are key.


\section{Acknowledgement}

We want to extend our gratitude to Mitsubishi Electric Co. Their invaluable insights and profound suggestions were crucial in laying the foundation of this project. We are also grateful for their financial support and for providing us with the opportunity that empowered us to realize this research.
   
\clearpage
 \bibliographystyle{elsarticle-num} 
 \bibliography{elsarticle-template-num}





\end{document}